%% file: robustcorr.tex
\documentstyle[jair,twoside,url,11pt,theapa]{article}

\jairheading{??}{2003}{?-??}{03/02}{09/03}
\ShortHeadings{Representation Dependence}
{Halpern \& Koller}
\firstpageno{1}

\input{defn}

\input{bghkmac}

\newcommand{\KLD}[3]{{\em KL}_{#1}({#2} \| {#3})}
\newcommand{\dists}{\Delta}
\renewcommand{\L}{{\cal L}}
\newcommand{\pr}{\mbox{\scriptsize\it pr\/}}

\newcommand{\LPr}{{\cal L}^{\pr}}

\newcommand{\intension}[2]{[\![ #1 ]\!]_{#2}}

\renewcommand{\dentails}{{\;|\!\!\!\sim\,}}
\newcommand{\dentailsn}{{\;|\!\!\!\sim}}

\renewcommand{\dentailssub}[1]{\dentails\hspace{-0.1em}_{
          \mbox{\scriptsize{\it #1}}}\,}

\newcommand{\red}{\mbox{\it red\/}}
\newcommand{\sparrow}{\mbox{\it sparrow\/}}
\newcommand{\blue}{\mbox{\it blue\/}}
\newcommand{\colorful}{\mbox{\it colorful\/}}
\newcommand{\lset}{}
\newcommand{\rset}{}
\newcommand{\noncolorful}{\mbox{\it colorless\/}}

\newcommand{\flyingbird}{\mbox{\it flying-bird\/}}
\renewcommand{\KBfly}{\KB^{\mbox{\scriptsize \it fly\/}}}
\newcommand{\me}{\mbox{\scriptsize\it me\/}}
\newcommand{\infp}{I}

\newcommand{\green}{\mbox{\it green\/}}

\renewcommand{\mid}{\:|\:}
\newcommand{\proj}{{\it proj}}

\begin{document}
\title{Representation Dependence in Probabilistic Inference}
\author{\name Joseph Y.\ Halpern \email halpern@cs.cornell.edu\\
\addr Cornell University, Computer Science Department\\
Ithaca, NY 14853\\
{\tt   http://www.cs.cornell.edu/home/halpern}
\AND
\name Daphne Koller \email koller@cs.stanford.edu\\
\addr Stanford University, Computer Science Department\\
Stanford, CA 94035\\
{\tt http://www.cs.stanford.edu/~koller}
}
\maketitle
\begin{abstract}
Non-deductive reasoning systems are often {\em representation
dependent}:~representing the same situation in two different ways
may cause such a system to return two different answers.  
Some have viewed this as a significant problem.  
For example, the {\em
 principle of maximum entropy\/} has been subjected to much criticism
due to its representation dependence.  There has, however, been almost
no work investigating representation dependence.  In this paper, we
formalize this notion and show that it is not a problem specific to
maximum entropy.  In fact, we show that any representation-independent
probabilistic inference procedure 
that ignores irrelevant information is essentially entailment, in a
precise sense.  Moreover, we show that representation independence is
incompatible with even a weak
default assumption of independence.
We then show that invariance under a
restricted class of representation changes can form a reasonable
compromise between representation independence and other desiderata,
and provide a construction of a family of inference procedures that
provides such restricted representation independence, using
relative entropy.
\end{abstract}

\section{Introduction}

It is well known that the way 
in which 
a problem is represented can have
a significant impact on the ease with which people solve it, and on the
complexity of an algorithm for solving it.  We are
interested in what is arguably an even more fundamental issue:
the extent to which the {\em answers\/} that we get depend on how our
input is represented.  
Here too, there is well known work, particularly by Tversky and
Kahneman (see, for example, \cite{KST82}), showing that the answers
given by people can 
vary significantly (and in systematic ways) depending on  how a question
is framed.  This 
phenomenon is often viewed as indicating a problem with human
information processing; the implicit assumption is that although people
do make mistakes of this sort, they shouldn't.  On the other hand, there
is a competing intuition that suggests that representation does (and {\em
should\/}) matter; representation dependence is just a natural
consequence of this fact. 

Here we consider one type of reasoning, probabilistic inference, and
examine the extent to which answers depend on the representation.
The issue of representation
dependence is of particular interest in this context because of the
interest in using probability for knowledge representation
(e.g.,~\cite{Pearl}) and because 
probabilistic inference
has been the source of many of the
concerns expressed regarding representation.
However, our approach should be applicable far more generally.
We begin by noting that the notion of ``probabilistic inference'' has two
quite different interpretations.  In one interpretation, which forms the
basis for the Bayesian paradigm, probabilitic inference consists basically
of conditioning:  We start out with a prior distribution over some event
space, and then condition on whatever observations are
obtained.  In the other interpretation, we are given only a set of
probabilistic assertions, and our goal is to reach conclusions about the 
probabilities of various events.  For most of this paper, we focus on the
latter interpretation, although we discuss the relationship to the
Bayesian approach in Section~\ref{sec:Bayesianapproach}.

Suppose that we have a procedure for making inferences from a probabilistic
knowledge base.  How sensitive is it to the way knowledge is
represented?
Consider the following examples, which use perhaps the
best-known non-deductive notion of probabilistic inference, {\em maximum
entropy\/} \cite{Jaynes}.%
\footnote{Although much of our discussion is motivated by the
representation-dependence problem encountered by maximum entropy,
an understanding of maximum entropy and how it works is not essential
for understanding our discussion.}
\xam\label{colorful}
Suppose that we have no information whatsoever
regarding whether an object is colorful.
What probability should we
assign to the proposition $\colorful$?  Symmetry arguments might suggest
$1/2$. Since we have no information, it seems that an object should be
just as likely to be colorful as non-colorful.  This
is also the conclusion reached by maximum entropy
provided that the language has only the proposition $\colorful$.
But now suppose we know about the colors red, blue, and green, and have
propositions corresponding to each of these colors.
Moreover, by 
$\colorful$ we actually mean $\red \lor \blue \lor \green$.  
In this
case, maximum entropy dictates that the probability of $\red \lor
\blue \lor \green$ is $7/8$.  Note that, in both cases, the only
conclusion that follows from our constraints is the trivial one: that
the probability of the query is somewhere between 0 and 1.
\exam

\xam\label{flyingbird}
Suppose that we are told that half of the birds fly.  There are two
reasonable ways to represent this information. One is to have
propositions $\bird$ and $\fly$, and use a knowledge base $\KBfly_1
\eqdef [\Pr(\fly \mid \bird) = 1/2]$.  A second might be to have as basic
predicates $\bird$ and $\flyingbird$, and use a knowledge base $\KBfly_2
\eqdef [(\flyingbird \rimp \bird) \land \Pr(\flyingbird \mid \bird) =
1/2]$.
Although the first representation may appear more natural, it seems
that both representations are intuitively 
adequate
insofar as representing the information that we have been given.  But
if we use an inference method such as maximum entropy, the first
representation leads us to infer $\Pr(\bird) = 1/2$, while the second
leads us to infer $\Pr(\bird) = 2/3$.
\exam

Examples such as these are the basis for the frequent criticisms of
maximum entropy on the grounds of representation dependence.  But
other than pointing out these examples, there has been little work on
this problem.  In fact, other than the work of 
Salmon~\citeyear{Salmon1,Salmon2}
and Paris \citeyear{Paris94},
there seems to have been no work on
formalizing the notion of representation dependence.
One might say that the consensus was: ``whatever representation
independence is, it is not a property enjoyed by maximum entropy.''
But are there any other inference procedures that have it?
In this paper we attempt to understand the notion of representation
dependence, and to study the extent to which it is achievable.

To study representation dependence, we must first understand
what we mean by a ``representation''.  The real world is complex.  In
any reasoning process, we must focus on certain details and ignore
others.  At a semantic level, the relevant distinctions are captured by
using a space $X$ of possible alternatives or states (possible worlds). 
In Example~\ref{colorful}, our first representation focused on the single
attribute $\colorful$.  In this case, we have only two states in the
state space, corresponding to $\colorful$ being true and false,
respectively.
The second representation, using $\red$, $\blue$, and $\green$, has a
richer state space.  Clearly, there are other distinctions that we
could make.  

We can also interpret a representation as a syntactic entity.  In this
case, we typically capture relevant
distinctions using some formal language.  For example, if we use
propositional logic as our basic knowledge representation language,
our choice of primitive propositions characterizes the distinctions
that we have chosen to make.  We can then take the states
to be truth assignments to these propositions.  Similarly,
if we use 
a probabilistic representation language such as 
{\em belief networks\/}~\cite{Pearl} as our knowledge
representation language, we must choose some set of relevant random
variables.  The states are then then possible assignments of values to
these variables.

What does it mean to shift from a representation (\ie state space)
$X$ to another representation $Y$?  
Roughly speaking, we want to capture at the level of the state space a
shift from, say, feet to meters.
Thus, in $X$ distances might be described in terms of feet
where in $Y$ they might be described in terms of meters.
We would expect there to be a constraint relating feet to meters.
This constraint would not give any extra information
about $X$; it would just relate worlds in $X$ to worlds in $Y$.  
Thus, we first attempt to capture representation independence somewhat 
indirectly, by requiring that adding constraints relating $X$ to
$Y$ that place no constraints on $X$ itself should not result in
different conclusions about $X$.  The resulting notion, called {\em
robustness}, turns out to be surprisingly strong.  We can show that
every robust inference procedure must behave essentially like logical
entailment.   

We then try to define representation independence 
more directly, by using a mapping $f$ from one
representation to another.  For example, $f$ could map a world where an
individual is 6 feet tall to the
corresponding world where the individual is 1.83 meters tall.
Some obvious constraints on $f$ are necessary to ensure that it
corresponds to our intuition of a representation shift.  
We can then define a
{\em representation-independent\/} inference procedure as one that 
preserves inferences under every legitimate mapping $f$; \ie  for any
$\KB$ and $\theta$, $\KB \dentails \theta$ iff $f(\KB) \dentails f(\theta)$.  

This definition turns out to be somewhat more reasonable than our first
attempt, in that there exist nontrivial representation-independent
inference procedures.  However, 
it is still 
a strong notion.  In particular, any
representation-independent inference procedure must act essentially like
logical entailment for a knowledge base with only 
{\em objective\/} information (i.e., essentially non-probabilistic
information).  
Moreover, we show that representation independence is incompatible 
with even the simplest default
assumption of independence.  Even if we are told nothing about the 
propositions $p$ and $q$, representation independence does not allow us to
jump to the conclusion that $p$ and $q$ are independent. 

\commentout{
For us, 
such a shift maps subsets of $X$ to subsets of $Y$.  Thus, for example,
the state where $\colorful$ holds can be mapped to the set of
states where either $\red$, $\blue$, or $\green$ holds.
We capture the notion of representation shift formally by
{\em embeddings}.
An {\em $X$-$Y$ embedding\/} $f$ maps subsets of $X$ to
subsets of $Y$ in a way that preserves complementation and
union.
An embedding is just the semantic version of the standard logical notion
of {\em interpretation\/} \cite[pp.~157--162]{Enderton}.
Essentially, an interpretation maps formulas in
a vocabulary $\Phi$ to formulas in a  different vocabulary $\Psi$ by
mapping the primitive propositions in $\Phi$ (e.g., $\colorful$) to
formulas over $\Psi$ (e.g., $\red \lor \blue \lor \green$) and then
extending to complex formulas in the obvious way.  The representation
shift in Example~\ref{flyingbird} can also be captured in terms of an
interpretation, this one taking $\flyingbird$ to $\fly \land \bird$.

Of course, not all embeddings count as legitimate representation
shifts.  For example, consider an embedding $f$ defined in terms of
an interpretation that maps both the propositions $p$ and
$q$ to the proposition $r$.  Then the process of changing representations
using $f$ gives us the information that $p$ and $q$ are equivalent,
information that we might not have had originally.  
We define a notion of a {\em faithful embedding\/} which is guaranteed
to give us no additional information.
An inference procedure allows us to take some information about the
world, and derive additional conclusions.  In the case of
probabilistic reasoning, both our knowledge base and our conclusions
take the form of statements about the probabilities of various events.
That is, we are interested in 
assertions,
such as $\Pr(\fly \mid \bird) = 1/2$, that
place constraints on 
probability measures on $X$.
Thus, a {\em probabilistic inference procedure\/} $\dentails$
takes a knowledge base consisting of assertions about probabilities
and uses it to reach conclusions about 
other such assertions.
A representation shift $f$ from $X$ to $Y$ induces a corresponding shift
$f^*$ {f}rom information about $X$ to information about $Y$.
For example, if the embedding $f$ maps
$\colorful$ to $\red \lor \blue \lor \green$, then $f^*(\Pr(\colorful)
\geq 2/3)$ is $\Pr(\red \lor \blue \lor \green) \geq 2/3$.
We say that an inference procedure $\dentails$ is
{\em invariant under $f$\/} if $f$ does not change the conclusions that we
make; that is, if for any $\KB$ and $\theta$, $\KB \dentails \theta$ iff
$f(\KB) \dentails f(\theta)$.  
We can now define $\dentails$ to be {\em
representation independent\/} if it is invariant under all 
faithful 
embeddings.  This 
definition
captures the intuition that $\dentails$ gives us the same answers no
matter how we shift representations.  

At first glance, representation independence seems like a reasonable
desideratum.  However, as we show in Section~\ref{sec:rep-ind}, it has
some rather unfortunate consequences.  In particular, we show that
any representation-independent inference procedure must act
essentially like logical entailment for a knowledge base with
only non-probabilistic information.  
Yet more troubling is the fact that
representation independence is completely incompatible with even the
simplest default assumption of independence: Even if we are told nothing
about the basic propositions $p$ and $q$, representation independence does
not allow us to jump to the conclusion that $p$ and $q$ are independent.
}

These results suggest that 
if we want inference procedures that are capable of jumping to
nontrivial conclusions, then we must accept at least some degree of
representation dependence.
They add support to the claim that the choice of language does
carry a great deal of information, and that complete representation
independence is too much to expect.
On a more positive note, we show that
we can use the intuition that the choice of
language carries information to get limited forms of representation
independence. 
The idea is that the language should put further constraints on what
counts as an ``appropriate'' representation shift.
For example, suppose that
certain propositions represent colors while others represent
birds.  While we may be willing to transform $\colorful$ to $\red \lor \blue
\lor \green$, we may not be willing to transform $\red$ to $\sparrow$.
There is no reason to demand that an inference
procedure behave the same way if we suddenly shift to a wildly
inappropriate representation, where the symbols mean something
completely different.
We provide a general approach to constructing
inference procedures that are invariant under a specific class of
representation shifts.
\commentout{
We assume that the user starts with some set of
initial {\em prior probability
measures\/} that characterize her beliefs in the absence of information.
We show that if these priors are chosen appropriately,
so that they are invariant under the class of 
embeddings of interest,
then we can ``bootstrap'' up to obtain a general inference procedure
that is invariant under the same class of embeddings.
The primary tool used in the bootstrapping process is {\em
cross-entropy\/}~\cite{X.entropy}, a well-known 
generalization of probabilistic conditioning.

This result can be used in a number of ways.  For example, it shows us
how to construct an inference procedure that is invariant under a given
set of embeddings:  we simply choose a class of priors appropriately.
} %
This construction allows us to combine some degree of representation
independence with certain non-deductive properties that we want of our
inference procedure.  
In particular, we present 
an inference method that supports a default assumption of independence,
and yet is invariant under a natural class of 
representation shifts.

The rest of this paper is organized as follows.  In
Section~\ref{procedures}, we define probabilistic inference procedures 
and characterize them.  In Section~\ref{sec:robust}, we define robust
inference procedures and show that every robust inference procedure is
essentially entailment.  In Section~\ref{sec:rep-ind}, we define
representation independence, and show that representation independence
is a very strong requirement.  In particular, we show that a
representation-independent inference procedure essentially acts like
logical entailment on objective knowledge bases and that representation
independence is incompatible with a default assumption of independence.
Section~\ref{discussion} contains some general discussion of the notion
of representation independence and how reasonable it is to assume that 
the choice of language should affect inference.  While it may indeed
seem reasonable to assume that the choice of language should affect
inference, we point out that this assumption has some consequences that
some might view as unfortunate.
In Section~\ref{limited}, we
discuss how limited forms of representation independence can be
achieved.  We discuss related work in Section~\ref{relatedwork}, and
conclude in Section~\ref{conclusions}.

\section{Probabilistic Inference}\label{procedures}
We begin by defining probabilistic inference procedures.  
As we discussed in the introduction, there are two quite different ways in
which this term is used.  In one, we are given a prior distribution over
some probability space; our ``knowledge'' then typically consists of
events in that space, which can be used to condition that distribution and
obtain a posterior.  In the other, which is the focus of our work, a
probabilistic inference procedure
takes as input a probabilistic knowledge base and returns a probabilistic
conclusion.  

We take both the knowledge base and the conclusion to be 
assertions about the probabilities of events in some 
{\em measurable space\/} $(X,\F_X)$, where a measurable space consists
of a set $X$ and an algebra $\F_X$ of subsets of
$X$ (that is, $\F_X$ is a set of subsets of $X$ closed under union and 
complementation, containing $X$ itself).%
\footnote{If $X$ is infinite, we may want
to consider countably-additive probability measures and take $\F_X$ to
be closed under countable unions.  This issue does not play  significant
role in this paper.  For simplicity, we restrict to finite additivity
and require only that $\F_X$ be closed under finite unions.}
Formally, these assertions can be viewed as statements about (or
constraints on) probability measures on $(X,\F_X)$.  For example,
if $S \in \F_X$, a statement $\Pr(S) \geq 2/3$
holds only for distributions where $S$ has
probability at least $2/3$.  Therefore, if $\dists_{(X,\F_X)}$ is the
set of all 
probability measures on $(X,\F_X)$ (that is, all probability measures
with domain $\F_X$), we can view a knowledge base as a set
of constraints on $\dists_{(X,\F_X)}$.  When $\F_X$ is clear from context,
we often omit it from the notation, writing $\dists_X$ rather than
$\dists_{(X,\F_X)}$.  

We place very few restrictions on the language used to express the
constraints.  
We assume that it 
includes
assertions of the form $\Pr(S) \ge \alpha$
for all subsets $S \in \F_X$ and rational $\alpha \in [0,1]$, and that 
it is closed under
conjunction and negation, so that if $\KB$ and $\KB'$ are knowledge
bases expressing constraints, then so are $\KB \land \KB'$ and $\neg
\KB$.
(However, the langauge could include many assertions besides those
obtained by starting with assertions of the form $\Pr(S) \ge \alpha$ and
closing off under conjunction and negation.)
Since the language puts constraints on probability measures, we cannot
directly say that $S \in \F_X$ must hold.  The closest approximation in the
language is the assertion $\Pr(S) = 1$.  Thus, we call such constraints {\em
objective}.  
A knowledge base consisting of only objective constraints is called an
{\em objective knowledge base}.  
Since $\Pr(T_1) = 1 \land \Pr(T_2) = 1$ is equivalent to $\Pr(T_1 \inter
T_2) = 1$, without loss of generality, an objective knowledge base
consists of a single constraint of the form $\Pr(T) = 1$.
Given a knowledge base $\KB$ placing constraints on $\dists_X$,
we write $\mu \sat \KB$ if $\mu$ is a measure in $\dists_X$ that
satisfies the constraints in $\KB$.  We use
$\intension{\KB}{X}$ to denote all the measures satisfying these
constraints.  

In practice, our knowledge is typically represented syntactically,
using some logical language to describe the possible 
states. 
Typical languages include
propositional logic, first-order logic, or a language describing the
values for some set of random variables.  In general, a base logic
$\L$ defines a set of formulas $\L(\Phi)$ for a given vocabulary
$\Phi$.  In propositional logic, the vocabulary $\Phi$ is simply a set
of propositional symbols.  In probability theory, the vocabulary can
consist of a set of random variables.  In first-order logic, the
vocabulary is a set of constant symbols, function symbols, and
predicate symbols.
To facilitate comparison between vocabularies,
we assume that for each base
logic all the vocabularies are finite subsets of one fixed infinite
vocabulary $\Phi^*$.

When working with a language, we assume that 
each state in the state space defines an interpretation for the
symbols in $\Phi$ and hence for the formulas in $\L(\Phi)$. In the case
of propositional logic, we thus assume that we can associate with each
state a truth assignment to the  
primitive propositions in $\Phi$.  For first-order logic, we assume that we can
associate with each state a domain and an interpretation of the symbols
in $\Phi$.  In the probabilistic setting, we assume that we can
associate with each state an assignment of values to the 
random variables.
It is often convenient to assume that the state space is in fact 
some subset $W$ of $\W(\Phi)$, the set of all interpretations for (or
assignments to)  the vocabulary $\Phi$.
Note that the truth of any formula $\phi$ in
$\L(\Phi)$ is determined by a state.  If $\phi$ is true in some state
$w$, we write $w \sat \phi$.

The {\em probabilistic extension\/} $\LPr(\Phi)$ of a
base logic $\L(\Phi)$ is simply the set of probability formulas over
$\L(\Phi)$.  Formally, for each $\phi \in \L(\Phi)$, $\Pr(\phi)$ is a
numeric term. The formulas in $\LPr(\Phi)$ are defined to be all the
Boolean combinations of arithmetic expressions involving numeric terms.
For example, $\Pr(\fly \mid \bird) \geq
1/2$ is a formula in $\LPr(\{\fly,\bird\})$
(where we interpret a conditional probability expression
$\Pr(\phi \mid \psi)$ as $\Pr(\phi \land \psi)/\Pr(\psi)$ and then
multiply to clear the denominator).
By analogy with constraints, a formula of the form $\Pr(\phi) = 1$ is
called an {\em objective formula}.

Given a set $W \subseteq \W(\Phi)$, assume that $\F_W$ is the algebra
consisting of all sets  of the form 
$\intension{\phi}{W} = \{w: w \sat \phi\}$, for $\phi \in \L(\Phi)$.
(In the case of 
propositional logic, where $\Phi$ consists of a finite set of primitive
propositions, $\F_W = 2^W$.  In the case of first-order logic, not all
sets are necessarily definable by formulas, so $\F_W$ may be a strict
subset of $2^W$.)  Let $\mu$ be a probability measure on
$(W,\F_W)$.  We can then ascribe semantics to $\LPr(\Phi)$ 
in the probability space
$(W,\F_W,\mu)$ in a straightforward way.  In particular, we interpret the
numeric term $\Pr(\phi)$ as 
$\mu(\{w \in W \, :\, w \sat \phi\})$.
Since a formula $\phi \in \L(\Phi)$ describes an event in the space $W$,
a formula $\theta$ in $\LPr(\Phi)$ is clearly a constraint on
measures on $W$.  We write $\mu \sat \theta$ if the measure
$\mu \in \dists_W$ satisfies the formula $\theta$.
A syntactic knowledge base $\KB \in \LPr(\Phi)$ can be viewed as a
constraint on 
$\dists_W$ in an obvious way.  Formally, 
$\KB$ represents the set of probability measures $\intension{\KB}{\Phi}
\subseteq \dists_W$,
which consists of all measures $\mu$ on $W$ such that
$\mu \sat \KB$.

We say that $\KB$ (whether syntactic or semantic) is {\em consistent\/} if
$\intension{\KB}{X} \ne \emptyset$, \ie if the constraints are
satisfiable. Finally, we say that $\KB$ {\em entails\/} $\theta$ (where
$\theta$ is another set of constraints on $\dists_X$), written $\KB \sat_X
\theta$, if $\intension{\KB}{X} \subseteq \intension{\theta}{X}$, \ie if
every measure that satisfies $\KB$ also satisfies $\theta$.  We write
$\sat_X \theta$ if $\theta$ is satisfied by every measure in
$\dists_X$.  We omit the subscript $X$ from $\sat$ if it is clear from
context.

Entailment is well-known to be to be a very weak method of drawing
conclusions from a knowledge base,
in particular with respect to its treatment of
irrelevant information.  Consider the knowledge base
consisting only of the constraint
$\Pr(\fly \mid \bird) \geq 0.9$.  Even
though we know nothing to suggest that $\red$ is at all relevant,
entailment will not allow us to reach any nontrivial conclusion about
$\Pr(\fly \mid \bird \land \red)$.  

One way to get
more powerful conclusions is to consider, not all the measures that
satisfy $\KB$, but a subset of them.   Intuitively, given a knowledge
base $\KB$, an inference procedure picks a subset of the measures
satisfying $\KB$, and infers $\theta$ if $\theta$ holds in this subset.
Clearly, more conclusions hold for every measure in the subset
than hold for every measure in the entire set.

\dfn An {\em $(X,\F_X)$-inference procedure\/} is a 
partial
function $\infp: 2^{\dists_{(X,\F_X)}} \mapsto 2^{\dists_{(X,\F_X)}}$
such that
$\infp(A) 
\subseteq A$ for $A \subseteq \dists_{(X,\F_X)}$ and $\infp(A) =
\emptyset$ iff $A = \emptyset$
for all $A \in  2^{\dists_{(X,\F_X)}}$ in the domain of $I$ (i.e., for
all $A$ for which $I$ is defined).
We write $\KB
\dentailsn_{\infp}\, \theta$ if
$\infp(\intension{\KB}{X}) \subseteq \intension{\theta}{X}$. 
\edfn
When $\F_X$ is clear from context or irrelevant, we often speak of
$X$-inference procedures.
We remark that Paris \citeyear{Paris94} considers what he calls {\em
inference processes}.  These are just inference procedures as we have
defined them that, given a a set $A$ of probability measures, return a
unique probability measure in $A$ (rather than an arbitrary subset of
$A$).  Paris gives a number of examples of inference processes.
He also considers various properties that an inference process might
have.  Some of these are closely related to various properties of
representation indepedence that we consider.  We discuss Paris's work in
Section~\ref{relatedwork}.

Entailment is the $X$-inference procedure 
defined on all sets 
determined by taking $\infp$
to be the identity.  Maximum entropy is also an inference procedure in
this sense.

\dfn\label{maxent} Given a probability measure $\mu$ on a finite space
$X$ (where all sets are measurable), 
its {\em entropy\/} $H(\mu)$ is defined as
$- \sum_{x \in X} \mu(x) \log \mu(x)$.
(The $\log$ is taken to the base 2 here.)
Given a set $A$ of measures in $\dists_X$,
let $\infp^{\me}_X(A)$
consist of the measures in $A$ that have the highest entropy
if there are measures in $A$ whose entropy is at least as high as that of any
measure in $A$; if there are no such measures, $\inf^{\me}(A)$ is undefined.
\edfn

It is easy to see that $\inf^{\me}(A)$ is defined if $A$ is closed (in
the topological sense; i.e., if $\mu_n$ is a sequence of probability
measures in $A$ and $\mu_n$ converges to $\mu$, then $\mu \in A$).
Thus, we could take the domain of $\inf^{me}_X$ to consist only of the
closed sets of measures in $\dists_X$.  
There are also open sets $A$ for which $\inf^{\me}(A)$ is defined,
although it is not defined for all open sets $A$.  For example,
suppose $X = \{x_1,x_2\}$ and let $A = \{\mu: \mu(x_1) < 1/2\}$.  
Let $\mu_0$ be such that $\mu_0(x_1) = 1/2$.  It is easy to check that
$H(\mu_0) = 1$, and $H(\mu) < 1$ for $\mu \in A$.  However, for all
$\epsilon$, there is some $\mu \in A$ such that $H(\mu) > 1 -
\epsilon$.  It follows that there is no measure in $A$ whose entropy is
higher than that of any other measure in $A$, so $\inf^{\me}(A)$ is
undefined.  On the other hand, if $A' = \{\mu: \mu(x_1) < 2/3\}$, then
there is a measure whose entropy is maximum in the open set $A'$, namely
the measure $\mu_0$.

There are, of course, many inference procedures besides entailment and
maximum entropy that can be defined on a measurable space.  
In fact, as the following proposition shows, any binary relation
$\dentails$ satisfying certain reasonable properties is an inference
procedure of this type.

\pro\label{choosedists.side1}
If $\infp$ is an $X$-inference procedure then the following
properties hold for every $\KB,\KB',\theta,\psi$ over $X$
such that $\KB$ is in the domain of $\infp$.
\begin{itemize}
\item
{\em Reflexivity:} $\KB \dentailsn_\infp\, \KB$.
\item {\em Left Logical Equivalence:}
if $\KB$ is logically equivalent to $\KB'$, \ie
if $\sat \KB \dimp \KB'$, then for every $\theta$ 
$\KB \dentailsn_\infp\, \theta$ iff $\KB' \dentailsn_\infp\, \theta$.
\item {\em Right Weakening}:
if $\KB \dentailsn_{\infp}\,
\theta$ and $\sat \theta \rimp \psi$ then $\KB
\dentailsn_\infp\, \psi$.
\item {\em And:}  
if $\KB \dentailsn_\infp\, \theta$ and $\KB
\dentailsn_\infp\, \psi$, then
$\KB \dentailsn_\infp\, \theta \land \psi$.
\item {\em Consistency\/}: if $\KB$ is consistent then $\KB
\mbox{$\;|\!\!\!\not\sim$}_\infp\, \false$. \ \ \bbox
\end{itemize}
\epro
\prf Straightforward from the definitions.  \eprf

Interestingly, these properties are commonly viewed as part of a core
of reasonable properties for a nonmonotonic inference relation
\cite{KLM}.

We would like to also prove a converse, showing that any relation
$\dentailsn$ over probabilistic constraints on some space $X$ that
satisfies the five properties above must have the form
$\dentailsn_{I_X}$.  This is not quite true, as the following example shows.
\xam Fix a measurable space $(X,\F_X)$.  Let the language consist of all
(finite) Boolean combination of statements of the form $\Pr(S) \ge
\alpha$, where $S \in \F_X$.  Now fix one nonempty strict subset $S_0$ of
$X$, and let $\phi_n$ be the statement $\Pr(S_0) \le 1/n$.  Define an inference
procedure $\dentailsn$ as follows.  If $\KB$ is not equivalent to
$\true$ (i.e, if $\intension{\KB}{X} \ne \dists_X$), then 
$\KB \dentails \theta$ iff $\KB \sat \theta$. On the other hand,
$\true \dentails \theta$ iff $\phi_n \sat \theta$ for all sufficiently
large $n$. That is, $\true \dentails \theta$ if there exists an $N$ such that
for all $n \ge N$, we have $\phi_n \sat \theta$.  It is easy to
check that all five properties in Proposition~\ref{choosedists.side1}
hold for $\dentailsn$.  However, $\dentailsn$ is not $\dentailsn_{I}$
for an $X$-inference procedure $I$.  For suppose it were.  Note that
$\phi_n \sat \phi_m$ for all $n \ge m$, so $\true \dentails \phi_m$ for
all $m$.  Thus, we 
must have $I(\dists_X) \subseteq \intension{\phi_n}{X}$ for all $n$.
It follows that $I_X(\dists_X) \subseteq \intension{\Pr(S_0) = 0}{X}$,
and so $\true \dentailsn_{I} \Pr(S_0) = 0$.  However,
$\phi_n \not \sat \Pr(S) = 0$ for any $n$, so we do not have
$\true \dentails \Pr(S) = 0$.  This contradicts the assumption that
$\dentails = \dentailsn_{I}$.
\exam

Essentially what we need to get the converse to
Proposition~\ref{choosedists.side1} is an infinitary version of the And
Rule,
which would say that if $\KB \dentailsn_{\infp\,} \theta_i$ for all $i$,
then $\KB \dentailsn_\infp\, \band_i \theta_i$.  
If the language were closed under infinite conjunctions, 
then this rule would in fact be just what we need.  Since we have not
assumed that the language is closed under infinite conjunctions, we
use a variant of this rule.

\begin{itemize}
\item {\em Infinitary And:} For any set $\Sigma$ of statements,
if $\KB \dentailsn_\infp\, \theta$ for all $\theta \in \Sigma$ and
$\Sigma \sat \psi$, then $\KB
\dentailsn_\infp\, \psi$. 
\end{itemize}

\pro\label{choosedists.side2}
Let $\dentails$ be a relation over probabilistic constraints 
on $X$
for which the properties Reflexivity, Left Logical Equivalence,
Right Weakening, Infinitary And, and Consistency hold
for all $\KB$ in the domain of $\dentails$.  (That is, if $\KB$ is in
the domain of $\dentails$ in that $\KB \dentails \theta$ for some
$\theta$, then $\KB \dentails \KB$, and so on.) 
Then $\dentails$ is $\dentailsn_{\infp}$ for some $X$-inference procedure 
$\infp$.
\epro
\prf See Appendix~\ref{sec2prfs}. \eprf

We are typically interested not just in an inference procedure defined
on one space $X$, but in a family of related inference procedures,
defined on a number of spaces.  For example, entailment is an
inference procedure that is defined on all spaces $X$; maximum entropy
is defined on all finite measurable spaces $(X,2^X)$.  

\dfn If $\X$ is a set of measurable spaces, an {\em $\X$-inference
procedure\/} is a set $\{I_{(X,\F_X)}: (X,\F_X) \in  
\X\}$, where $I_{(X,\F_X)}$ is an $(X,\F_X)$-inference procedure for
$(X,\F_X) \in \X$. 
\edfn

We sometimes talk about an $\X$-inference procedure $I$, and write $\KB
\dentailsn_{\infp}\, \theta$ when $(X,\F_X) \in \X$ is clear from context. 
However, it should be stressed
that, formally, an $\X$-inference procedure is a really a set of inference
procedures (typically related in some natural way).  

Clearly entailment is an $\X$-inference procedure for any $\X$, where
$\infp_X$ is simply the identity function for $X \in \X$.  
If $\X$ consists of finite measurable spaces where all sets are
measurable, then maximum entropy is an $\X$-inference procedure.  
We typically denote this inference procedure
$\dentailssub{me}$. 
Thus,
$\KB \dentailssub{me} \theta$ if $\theta$ holds for all the
probability measures of maximum entropy satisfying $\KB$.

{\bf Important assumptions:}  For the remainder of this paper, we deal
only with $\X$-inference procedures $I$ for which $\X$ 
satisfies two richness assumptions.  These assumptions hold for all
the standard inference procedures that have been considered.
\begin{itemize}
\item We assume that $\X$ is {\em closed under crossproducts}, so that
if $(X,\F_X), (Y,\F_Y) \in \X$, then $(X\times Y, \F_{X \times Y}) \in
\X$, where 
$\F_{X \times Y}$ is the algebra formed by taking finite unions of
disjoint sets of the form $S \times T$, for $S \in \F_X$ and $T \in
\F_Y$.  It is easy to see that this is an algebra, since $\overline{S
\times T} = \overline{S} \times T \union S \times \overline{T} \union
\overline{S} \times \overline{T}$ and $(S \times T) \inter (S' \times
T') = (S \inter S') \times (T \inter T')$
(from which it also follows that any
union of such sets can be written as a disjoint union).  Note that if
$X$ and $Y$ are finite sets, $\F_X = 2^X$, and $\F_Y = 2^Y$, then $\F_X
\times \F_Y = 2^{X \times Y}$.   
As we shall see, having $(X \times Y, \F_{X \times Y}) \in \X$ if each
of $(X,\F_X)$ and $(Y,\F_Y)$ is in 
$\X$ allows us to relate constraints on $X$ to constraints on
$Y$ in a natural way.  
\item We assume that
$\X$ contain sets of all finite cardinalities; more precisely, for all
$n \ge 2$, there exists a set $(X,\F_X) \in \X$ such that $|X| = n$ and
$\F_X = 2^X$.  This assumption is not actually needed for any of our
results, since the assumption that $\X$ is closed under crossproducts
already implies that, for any finite $n$, there exists a measurable
space $(X,\F_X) \in \X$ such that $|X| \ge n$; this already suffices to
prove all the results of the paper.  However, assuming that $\X$ has
sets of all cardinalities does make the proofs easier.
\end{itemize}

We also want the domain of $I$ to satisfy certain assumptions, but we
defer stating these assumptions until we have introduced some additional
definitions and notation. 

\section{Robustness}\label{sec:robust}

In order to define robustness to representation shifts, we must first define
the notion of a representation shift.  Our first attempt at this definition
is based on the idea of using constraints that specify the relationship
between the two vocabularies.  For example, in Example~\ref{colorful}, we
might have $X = \{\colorful, \noncolorful\}$ and $Y =
\{\red,\blue,\green,\noncolorful\}$.  We can specify the relationship
between $X$ and $Y$ via a constraint that asserts that $\colorful \dimp
(\red \lor \blue \lor \green)$.  

Of course, not every constraint is a legitimate mapping between
representations.  For example, a formula that asserted $\neg \colorful$ is
obviously not a legitimate representation shift.  At a minimum, we must
assume that the constraint does not give 
any additional information about $X$ as far as logical inference goes.
At a syntactic level, we can use the following definition.  Given a
knowledge base $\KB \in \LPr(\Phi)$, we say that 
$\psi \in \LPr(\Phi \union \Phi')$ is {\em
$\Phi$-conservative\/} over $\KB$ if, for all formulas $\phi \in
\LPr(\Phi)$, we have $\KB \sat \phi$ iff $\KB \land \psi \sat \phi$.  
Thus, adding $\psi$ to the knowledge base does not permit any additional
logical inferences in the vocabulary $\Phi$.
An inference procedure $I$ is {\em robust\/} if it
is unaffected by conservative extensions; that is, if $\KB, \phi \in
\LPr(\Phi)$, then $\KB \dentailsn_{I\,} \phi$ iff $\KB \land \psi
\dentailsn_{I\,} \phi$ 
for all $\psi$ that are $\Phi$-conservative over $\KB$.
Roughly speaking, this says that getting new information that is
uninformative as far as logical inference goes does not affect
default conclusions. 

The formal definition of robustness, which uses semantic rather than
syntactic concepts, extends these intuitions to arbitrary constraints on
measures (not just ones that can be expressed in the language $\LPr$).

\dfn
For $\mu \in \dists_{X_1 \times \ldots \times  X_n}$, define $\mu_{X_i} \in
\dists_{X_i}$  
by taking $\mu_{X_i}(A) = \mu(X_1 \times \cdots \times X_{i-1} \times A
\times X_{i+1} \times \cdots \times X_n)$.  
A constraint $\phi$ on $\dists_{X_i}$ can be viewed as a
constraint on $\dists_{X_1 \times \ldots \times X_n}$ by taking
$\intension{\phi}{X_1 \times \ldots \times X_n}
= \{\mu \in \dists_{X_1  \times \ldots \times X_n}: \mu_{X_i}\sat
\phi\}$.  We frequently 
identify constraints on $X_i$ with constraints on $X_1 \times \ldots
\times X_n$ in this way. 
For $B \subseteq \dists_{X_1 \times \cdots \times X_n}$, define $\proj_{X_i}(B) =
\{\mu_{X_i}: \mu \in B\}$.
A constraint $\psi$ on $\dists_{X_1 \times \cdots \times X_n}$ is said to be
{\em $X_i$-conservative over the constraint $\KB$ on $\dists_{X_i}$\/}
if $\proj_{X_i}(\intension{\KB \land \psi}{X_1 \times \cdots \times X_n}) =
\intension{\KB}{X_i}$. 
\edfn

To see that this definition generalizes 
the earlier language-oriented
definition, note that if $\phi$ and $\KB$ are
constraints on $\dists_X$ 
and $\psi$ is a constraint on $\dists_{X \times Y}$, 
 then $\KB
\land \psi \sat \phi$ iff $\proj_1(\intension{\KB \land \psi}{X \times Y})
\subseteq \intension{\phi}{X}$, while $\KB \sat \phi$ iff $\intension{\KB}{X}
\subseteq \intension{\phi}{X}$.

\dfn $\{I_X: X \in \X\}$ is a {\em robust\/} $\X$-inference procedure if
for all spaces $X, Y \in \X$,  
constraints $\KB$ and $\phi$ on $\dists_X$, and constraints $\psi$
on $\dists_{X \times Y}$ that are $X$-conservative over $\KB$, we have
$\KB \dentailsn_{I_X} \phi$ iff $\KB \land \psi \dentailsn_{I_{X \times
Y\,}} \phi$. 
(Note that this definition implicitly assumes that $X \times Y \in \X$
if $X, Y \in \X$, an assumption we made explicit earlier.)
\edfn

At first glance, robustness might seem like a reasonable desideratum. 
After all, why should 
adding a constraint on $\dists_{X \times Y}$ that places no restrictions
on $\dists_X$ change the conclusions that we might reach about $X$?
Unfortunately, it 
turns out that this definition is deceptively strong, and disallows any
``interesting'' inference procedures.  In particular, one property we may
hope for in an inference procedure is to draw nontrivial conclusions about
probabilities of events, that is, conclusions that do not follow from
entailment.  For example, maximum entropy (or any inference procedure based
on symmetry) will conclude $\Pr(p) = 1/2$ from the empty knowledge base.  
We can show that inference procedures that are robust do not really
allow much in the way of nontrivial conclusions about the probabilities of
events.

\dfn
An $(X,\F_X)$-inference procedure $\infp$ is {\em essentially
entailment for the knowledge base $\KB \subseteq \dists_{X}$\/} if
for all $S \in
\F_X$, if $\KB \dentailsn_\infp\, \alpha < \Pr(S) < \beta$ then
$\KB \sat \alpha \le \Pr(S) \le \beta$.  $\infp$ is {\em essentially
entailment for $X$\/} if it is essentially entailment for all knowledge
bases 
$\KB$ in the domain of $I_X$.
\edfn
Thus, when entailment lets us conclude $\Pr(S) \in [\alpha,\beta]$, an
inference procedure that is essentially entailment lets us draw only the
slightly stronger conclusion $\Pr(S) \in (\alpha,\beta)$.  
To prove this, we need to make three assumptions
about the domain of $I$.
(For other results, we need other assumptions about the
domain of $I$.)

\begin{itemize}
\item[DI1.] $\alpha \le \Pr(S) \le \beta$ is in the domain of
$I_{(X,\F_X)}$ for all $S \in \F_X$, $\alpha, \beta \in \IR$.
\item[DI2.] If $\KB$ is in the domain of $I_X$, then it is also in the
domain of $I_{X \times Y}$ (when $\KB$ is viewed as a constraint on
$\dists_{X \times Y}$.)
\item[DI3.] If $\KB_1$ and $\KB_2$ are in the domain of $I_X$, then so
is $\KB_1 \land \KB_2$.
\end{itemize}

Note that sets of the form $\alpha \le \Pr(S) \le \beta$ are closed
sets.  It certainly seems reasonable to require that such sets be in the
domain of an inference procedure; they correspond to the most basic
observations.  DI2 seems quite innocuous; as observed earlier, we do want
to be able to view constraints on $\dists_X$ as constraints on
$\dists_{X \times Y}$, and doing so should not prevent them from being
in the domain of $I$.  DI3 also seems to be a reasonable assumption, since if
$\KB_1$ and $\KB_2$ correspond to possible observations, we want to be
able to draw conclusions from combining the observations.  
DI3 holds if the domain of $I$ consists of closed sets.
But note that it does not hold for $I^{\me}$ if we take its domain to
consist of all sets that have a measure whose entropy is maximum.  For
example, if $X = \{x_1,x_2\}$, $A = \{\mu_0\} \union \{\mu:
\mu(x_1) > 3/4\}$, and $B = \{\mu:
\mu(x_1) \ge 2/3\}$, where $\mu_0(x_0) = 1/2$, then each of $A$ and $B$
have a measure whose entropy is maximum, but $A \inter
B$ does not have a measure whose entropy is maximum.

\thm\label{almosttrivial2}
If $\{I_X: X \in \X\}$ is a robust $\X$-inference procedure 
that satisfies DI1, DI2, and DI3,
then $I_X$
is essentially entailment for all $X \in \X$.
\ethm
\prf See Appendix~\ref{sec3prfs}. \eprf

It is possible to construct robust inference procedures that are almost but
not quite entailment, simply by ``strengthening'' some conclusions from
$\Pr(S) \in [\alpha,\beta]$ to $\Pr(S) \in (\alpha,\beta)$.  
Clearly, however, any robust inference procedure is extremely limited in its
ability to jump to conclusions.  In the next section, we look at a
definition that seems closer to the intuitive notion of
representation independence, and has somewhat more reasonable consequences.

\section{Representation Independence}\label{sec:rep-ind}

\subsection{Representation shifts}
If $X$ and $Y$ are two different representations
of the same phenomena 
then, intuitively, there should be a way of relating states in $X$ to
corresponding states in $Y$.
We want this correspondence to respect the logical structure of events.
Formally, we require that it be a homomorphism with respect to
complementation and intersection.
\dfn
An {\em $(X,\F_X)$-$(Y,\F_Y)$ embedding\/} $f$ is a function
$f: \F_X \mapsto \F_Y$ such that
$f(S \union T) = f(S) \union f(T)$
and $f(\overline{S}) = \overline{f(S)}$ for all $S, T
\in \F_X$.
\edfn
As elsewhere, we talk about $X$-$Y$ embeddings rather than
$(X,\F_X)$-$(Y,\F_Y)$ embeddings if $\F_X$ and $\F_Y$ do
not play a significant role.

Our goal is to consider the effect of a transformation on probabilistic
formulas.  Hence, we are interested in sets of states and their
probabilities. 
\dfn \label{dfn:correspond}
If $f$ is an $X$-$Y$ embedding, $\mu \in \dists_X$, and $\nu \in
\dists_Y$, then 
$\mu$ and $\nu$ {\em correspond under $f$}
if $\mu(S) = \nu(f(S))$ for all events $S \in \F_X$. We define 
a mapping $f^*: 2^{\dists_X} \mapsto 2^{\dists_Y}$ as follows.
We first define $f^*$ on singleton sets (except that, for convenience,
we write 
$f^*(\mu)$ rather than $f^*(\{\mu\})$ by taking
$f^*(\lset\mu\rset) = \{\nu \in \dists_Y: \nu(f(S)) = \mu(S) \mbox{ for all } S
\in \F_X\}$. 
Thus,
$f^*(\mu)$ consists of all measures in $\dists_Y$ that
correspond to $\mu$ under $f$.
If $\D$ is an arbitrary subset of $2^{\dists_X}$, define
$f^*(\D) = \union_{\mu \in \D} f^*(\lset\mu\rset)$ for
$\D \subseteq \dists_X$.
\edfn
If $\theta$ is 
a constraint on $\dists_X$
expressed in some language,
we typically write $f^*(\theta)$
rather than $f^*(\intension{\theta}{X})$. 
We implicitly assume that the language is such that the constraint
$f^*(\theta)$ is also expressible.  It is not hard to see that 
$f^*(\theta)$ is the constraint that results by replacing every set
$S\in \F_X$ that appears in $\theta$ by $f(S)$.

\xam\label{embed.color}
In Example~\ref{colorful}, we might have $X = \{\colorful,
\noncolorful\}$ and $Y = \{\red,\blue,$ $\green,\noncolorful\}$.  In this
case, we might have $f(\colorful) = \{\red,\blue,\green\}$ and
$f(\noncolorful) = \{\noncolorful\}$.  Consider the measure $\mu \in
\Delta_X$ such that $\mu(\colorful) = 0.7$ and $\mu(\noncolorful) = 0.3$.
Then $f^*(\mu)$ is the set of measures $\nu$ such that the total
probability assigned to the set of states $\{\red,\blue,\green\}$ by
$\nu$ is 0.7. Note that there are uncountably many such measures.
It is easy to check that if $\theta$ is a constraint on $\dists_X$ such
as $\Pr(\colorful) > 3/4$, then $f^*(\theta)$ is
$\Pr(\{\red,\blue,\green\}) > 3/4$. 
\exam

Embeddings can be viewed as the semantic analogue to
the 
syntactic
notion of {\em interpretation} defined in~\cite[pp.~157--162]{Enderton},
which has also been used in the recent literature on {\em abstraction\/}
\cite{GW92,NL94}.  Essentially, an interpretation maps formulas in
a vocabulary $\Phi$ to formulas in a  different vocabulary $\Psi$ by
mapping the primitive propositions in $\Phi$ (e.g., $\colorful$) to
formulas over $\Psi$ (e.g., $\red \lor \blue \lor \green$) and then
extending to complex formulas in the obvious way.  The representation
shift in Example~\ref{flyingbird} can also be captured in terms of an
interpretation, this one taking $\flyingbird$ to $\fly \land \bird$.

\dfn\label{embedding}
Let $\Phi$ and $\Psi$ be two vocabularies.  In the propositional case,
a {\em interpretation of $\Phi$ into $\Psi$\/} is a function $i$
that associates with every primitive proposition $p \in \Phi$ a formula
$i(p) \in \L(\Psi)$.
A more complex definition in the same spirit applies to first-order
vocabularies.  For example, if $R$ is a $k$-ary predicate, then
$i(R)$ is a formula with $k$ free variables.
\edfn
Given an interpretation $i$, we get a syntactic translation from
formulas in $\L(\Phi)$ to formulas in $\L(\Psi)$
using $i$ in the obvious way; for example,
$i((p \land \neg q) \lor r) = (i(p) \land \neg i(q)) \lor i(r)$ (see
\cite{Enderton} for the details).
Clearly an interpretation $i$ from $\Phi$ to $\Psi$ induces an
embedding $f$ from $W_1 \subseteq \W(\Phi)$ to $W_2 \subseteq \W(\Psi)$:~%
we map $\intension{\phi}{W_1}$
to $\intension{i(\phi)}{W_2}$.

Of course, not all embeddings count as legitimate representation
shifts.  For example, consider an embedding $f$ defined in terms of
an interpretation that maps both the propositions $p$ and
$q$ to the proposition $r$.  Then the process of changing representations
using $f$ gives us the information that $p$ and $q$ are equivalent,
information that we might not have had originally.  
Intuitively, $f$ gives us new information by telling us that a certain
situation---that where $p \land \neg q$ holds---is not possible.
\commentout{
As we argued in the introduction, not every embedding is appropriate
as a representation shift. The process of changing representation should
not give us any new information.  When does a shift give us new
information?  One obvious situation is when the shift makes impossible
something which we considered to be possible.  In our example from the
introduction, we had an interpretation $i$ with the property that $i(p) =
i(q)$ for two propositions $p$ and $q$.  Clearly, this interpretation
gives us new information: that $p$ and $q$ are equivalent.  
We want to specify conditions on embeddings that disallow ``informative'' 
embeddings such as this.  In this case, 
the associated 
}
More formally, 
the
embedding $f$ has the following undesirable property: it 
maps the set of states satisfying $p \land \neg q$ to the empty set.
This means a state where $p \land \neg q$ holds does not have
an analogue in the new representation.
We want to disallow such embeddings.
\dfn\label{faithfulKB}
An $X$-$Y$ embedding $f$ is {\em faithful\/}
if, for all $S, T \in \F_X$, we have $S \subseteq T$ iff
$f(S) \subseteq f(T)$.
\edfn

This definition has the desired consequence of 
disallowing embeddings that give 
new information as far as logical consequence goes.
\lem\label{char.faithful}
An $X$-$Y$ embedding $f$ is faithful if and only if for all constraints
$\KB$ and $\theta$, we have $\KB \sat \theta$ iff $f^*(\KB) \sat
f^*(\theta)$. 
\elem
\prf See Appendix~\ref{sec4prfs}. \eprf

It is clear that our embedding from Example~\ref{embed.color} is faithful:
$f(\colorful) = \{\red,\blue,\green\}$ and $f(\noncolorful) =
\noncolorful$.  
The following proposition gives further insight into faithful
embeddings.  
\pro\label{pro:correspond1}
Let $f$ be a faithful $X$-$Y$ embedding.  Then
the following statements are equivalent:  
\begin{itemize}
\item[(a)] $\mu$ and $\nu$ correspond
under $f$; 
\item[(b)] for all formulas $\theta$, $\mu \sat\theta$ iff $\nu \sat
f^*(\theta)$.
\end{itemize}
\epro
\prf See Appendix~\ref{sec4prfs}. \eprf

If the embedding $f$ is a ``reasonable'' representation shift, we would
like an inference procedure to return the same answers if we shift
representations using $f$.
\dfn\label{dfn.invariant}
If $X, Y \in \X$, then 
the $\X$-inference procedure $\{I_X: X \in \X\}$ is {\em invariant\/}
under the $X$-$Y$ embedding $f$ if
for all constraints
$\KB$ and $\theta$ on $\dists_X$, we have $\KB \dentailsn_{\infp_X\,}
\theta$ iff $f^*(\KB) 
\dentailsn_{\infp_Y}\, f^*(\theta)$. 
(Note that, in particular, this means that $\KB$ is in the domain of
$\dentails_{\infp_X}$ iff $f^*(\KB)$ is in the domain of
$\dentails_{\infp_Y}$.)
\edfn
\dfn
The $\X$-inference procedure $\{I_X: X \in \X\}$ is 
{\em representation independent\/} if
it is invariant under all faithful $X$-$Y$ embeddings for all $X, Y \in \X$.
\edfn

Since the
embedding for Example~\ref{embed.color} is faithful, any
representation-independent inference procedure would return the same answers
for $\Pr(\colorful)$ as for $\Pr(\red \lor \blue \lor \green)$.
The issue is somewhat more subtle for
Example~\ref{flyingbird}.  There, we would like to have an embedding $f$
generated by the interpretation $i(\flyingbird) = \fly \land \bird$ and
$i(\bird) = \bird$.  This is not a faithful embedding, since
$\flyingbird \rimp \bird$ is not a valid formula, while $i(\flyingbird
\rimp \bird)$ is $(\fly \land \bird) \rimp \bird$ which is valid.
Looking at this problem semantically, we see that the state corresponding
to the model where $\flyingbird \land \neg\bird$ holds is mapped to
$\emptyset$.  But this is clearly the source of the problem.
According to our linguistic intuitions for this domain, this is not a
``legitimate'' state.
Rather than considering all the states in
$\W(\{\flyingbird,\bird\})$,
it is perhaps more appropriate to consider the subset $X$ consisting
of the truth assignments characterized by the formulas
$\{\flyingbird
\land \bird, \neg \flyingbird \land \bird, \neg\flyingbird \land \neg
\bird\}$.  If we now use $i$ to embed $X$ into $\W(\{\fly,\bird\})$, the
resulting embedding is indeed faithful.  So, as for the previous example,
invariance under this embedding would guarantee that we get the same
answers under both representations.

\subsection{Representation-independent inference procedures}
Although the definition of representation independence seems natural, so did
the definition of robustness.  How do the two definitions relate to each
other?  First, we show that representation independence is a weaker notion
than robustness.   
For this result, we need to consider inference procedures that
satisfy two further assumptions.
\begin{itemize}
\item[DI4.] If $f$ is a faithful $X$-$Y$
embedding, then $\KB$ is in the domain of $I_X$ iff $f^*(\KB)$ is in the
domain of $I_Y$. 
\item[DI5.] If $\KB$ is in the domain of $I_{X \times Y}$, $f$ is a
faithful $X$-$Y$ embedding, and  
$\phi_1$ is a constraint on $\dists_X$, then 
$\KB \land (\phi_1 \dimp f^*(\phi_1))$ is 
in the domain of $I_{X \times Y}$.
\end{itemize}
DI4 is very natural and is satisfied by all the standard inference
procedures.   It is easy to check that if $\KB$ is closed iff 
$f^*(\KB)$ is closed.  
While DI5 may not appear so natural, it does hold for domains consisting
of closed sets, since it is not hard to check that $\phi \dimp 
f^*(\phi_1)$ is closed.  DI5 would follow from DI3 and the assumption
that $\phi \dimp f^*(\phi_1)$ is in the domain of $I_{X \times Y}$, but
it is actually weaker than the combination of these two assumptions.
In particular, it holds for the domain consisting of all sets on which
there is a measure of maximum entropy.

\thm\label{robust} 
If an $\X$-inference procedure is robust
that satisfies DI2, DI4, and DI5,
then it is
representation independent. 
\ethm
\prf See Appendix~\ref{sec4prfs}. \eprf

We have already shown that any robust inference procedure must be almost
trivial.  Are there any interesting representation-independent inference
procedures? 
As we shall see, the answer is mixed.  There are nontrivial
representation-independent inference procedures, but they are not very
interesting.

Our first result shows that representation independence, like robustness,
trivializes the inference procedure, but only for some knowledge bases.

\thm\label{almosttrivial1}
If $\{I_X: X \in \X\}$ is a representation-independent $\X$-inference
procedure 
then, for all $X \in \X$,  $I_X$ 
is essentially entailment for 
all objective knowledge bases in its domain.%
\footnote{In an earlier version of this paper \cite{HKijcai95}, we
claimed that any representation-independent inference procedure that
satisfied a minimal irrelevance property (implied by robustness, but not
equivalent to it) is essentially entailment for all knowledge bases.  
As Jaeger \citeyear{Jaeger96} shows,
an inference procedure along the lines of $I^1$ described below can be
constructed to show that this result is not correct.  We seem to need
the full strength of robustness.}
\ethm
\prf See Appendix~\ref{sec4prfs}. \eprf

\cor\label{almosttrivial3}
If $\{I_X: X \in \X\}$ is a representation-independent $\X$-inference
procedure, $\KB$ is objective, and
$\KB \dentailsn_{I\,} \alpha < \Pr(S) < 
\beta$ for some $\alpha \ge 0$ and $\beta \le 1$, then $\alpha = 0$ and
$\beta = 1$. \ecor

This result tells us that from an objective knowledge base
$\Pr(T) = 1$, we can reach only three possible conclusions about
a set $S$.  If $T \subseteq S$, then we can conclude that $\Pr(S) = 1$; if
$T \subseteq \overline{S}$, then we can conclude that $\Pr(S) = 0$;
otherwise, the {\em strongest\/} conclusion we can make about $\Pr(S)$
is that is somewhere between 0 and 1.

We can construct a representation-independent inference procedure that is
not entailment and has precisely this behavior
if we restrict attention to countable state spaces.
Suppose that $X$ is countable.
Given an objective knowledge base $\KB$ of the form $\Pr(T) = 1$, where
$T \in \F_X$, let $\KB^+$
consist of all formulas of the form $0 < \Pr(S) < 1$ for 
for all nonempty strict subsets $S$ of $T$ in $\F_X$.%
\footnote{The requirement that $X$ be countable is necessary here.  If $X$ is
uncountable and every singleton is in $\F_X$, then
$\KB^+$ is inconsistent if both $T$ and $\overline{T}$ are uncountable.
It is impossible that each of an uncountable collection of points has
positive measure.} 
We now define an $X$-inference procedure $\infp^0_X$ as follows: If $\KB$
is equivalent to an objective knowledge base, then
$\KB \dentailsn_{I^0\,} \phi$ if $\KB \land \KB^+ \sat \phi$;
if $\KB$ is not equivalent to an objective knowledge base,
then $\KB \dentailsn_{I^0\,} \phi$ if $\KB \sat \phi$.
It follows easily from Proposition~\ref{choosedists.side2} that
$\infp^0_X$ is indeed an inference procedure.
Moreover, it is not equivalent to the standard notion of
entailment; for example, we have $\true \dentailsn_{I^0\,} 0 < \Pr(p) < 1$,
while ${\not\sat} 0 < \Pr(p) < 1$.  Nevertheless, we can prove that
$\infp^0$ is representation independent.
\lem\label{I*}
Let $\X$ consist of only countable sets.
Then $\{\infp^0_X: X \in \X\}$ is a
representation-independent $\X$-inference procedure. 
\elem
\prf See Appendix~\ref{sec4prfs}. \eprf

While objective knowledge bases may not appear so interesting
if we restrict to propositional languages, for languages that
include first-order and statistical information they become quite
interesting.  Indeed, as shown in \cite{Bacchus,BGHKfull}, knowledge
bases with first-order and
(objective)
statistical information allow us to express
a great deal of the information that we naturally encounter.  
For example, we can express the fact that ``90\% of birds fly'' as an
objective statement about the number of flying birds in our domain relative
to the overall number of birds.
Of course, Theorem~\ref{almosttrivial1} applies immediately to such
knowledge bases.

Theorem~\ref{almosttrivial1} also implies that various inference procedures
cannot be representation independent.  In particular, since $\true
\dentailssub{me} \Pr(p) = 1/2$ for a primitive proposition $p$, it
follows that maximum entropy is not essentially entailment.  This
observation provides another proof that maximum entropy
is not representation independent.

It is consistent with Theorem~\ref{almosttrivial1} that there are
representation-independent 
inference procedures that are not almost entailment for probabilistic
knowledge bases.  For example, consider the $X$-inference procedure $I^1_X$
defined as follows.  Given $A \subseteq \dists_X$, if there exists some $S
\in  \F_X$ such that $A = \{\mu \in \dists_X : \mu(S) \ge 1/4\}$, then
$I^1_X(A) = \{\mu \in \dists_X: \mu(S) \ge 1/3\}$; otherwise, $I^1_X(A) =
A$.  Thus, $\Pr(S) \ge 1/4 \dentailsn_{I^1} \Pr(S) \ge 1/3$.  Clearly,
$I^1_X$ is not essentially entailment.  Yet, we can prove the following result.

\lem\label{I1} 
Suppose that $\X$ consists only of measure spaces of the form $(X,2^X)$,
where $X$ is finite.  Then  $\{I^1_X: X \in \X\}$ 
is a representation-independent $\X$-inference
procedure. \elem
\prf See Appendix~\ref{sec4prfs}. \eprf

Note that it follows from Theorem~\ref{almosttrivial2} that $I^1$ cannot be
robust.  Thus, we have shown that representation independence is a strictly
weaker notion than robustness.  

This example might lead us to believe that there are 
representation-independent
inference procedures
that are ``interesting'' for probabilistic knowledge bases.  However, as we
now show, 
a representation-independent inference procedure 
cannot satisfy one key desideratum:
the ability to conclude independence by default.
For example, an important feature of the
maximum-entropy approach to nonmononotic reasoning \cite{GMPfull} has
been its ability to ignore ``irrelevant'' information, by implicitly
assuming independence.  Of course, maximum entropy does not satisfy
representation independence.  Our result shows that no approach to
probabilistic reasoning can simultaneously assure representation
independence and a default assumption of independence.

We do not try to give a general notion of ``default
assumption of independence'' here, since we do not need it for our
result.  Rather, we give a minimal property that we would hope an
inference procedure might have, and show that this property is sufficient to 
preclude representation independence.   Syntactically, the property we want
is that if $\Phi$ and $\Psi$ are disjoint vocabularies, $\KB \in
\LPr(\Phi)$, $\phi \in \L(\Phi)$, and $\psi \in \L(\Psi)$, then
$\KB \dentailsn_\infp \Pr(\phi \land \psi) = \Pr(\phi) \times \Pr(\psi)$.

\dfn
An $\X$-inference procedure $\{I_X: X \in \X\}$ enforces {\em minimal default
independence\/} if, whenever $X$ and $Y$ are 
in $\X$, $\KB$ is a constraint on $\dists_X$ 
in the domain of $\dentailsn_{\infp_{X}}$,
$S \in \F_X$, and $T \in \F_Y$,
then
$\KB \dentailsn_{\infp_{X \times Y}} \Pr(S
\times T) = \Pr(S) \times \Pr(T)$.%
\footnote{Since we are working in the space $X \times Y$, $\KB$ should
be viewed as a constraint on $\dists_{X \times Y}$ here, $\Pr(S)$
should be understood as $\Pr(S \times Y)$, while $\Pr(T)$ should be
understood as $\Pr(X \times T)$.  Recall that, by assumption, $X \times
Y \in \X$.}
\edfn
This definition clearly generalizes the syntactic 
definition.

Clearly, entailment does not satisfy minimal default independence.  Maximum
entropy, however, does.  Indeed, a semantic property that implies minimal
default independence is used by Shore and Johnson~\citeyear{ShoreJohnson} as 
one of the axioms in an axiomatic characterization of maximum-entropy.

\thm\label{noindep}
If $\{I_X: X \in \X\}$ is an $\X$-inference procedure that 
enforces minimal default independence
and satisfies DI1,
then $I_X$ is not representation independent.
\ethm
\prf See Appendix \ref{sec4prfs}. \eprf

This result is very interesting as far as irrelevance is concerned.
We might hope that learning irrelevant information does not affect our
conclusions.  While we do not attempt to define irrelevance here,
certainly we would expect that if $\KB'$ is in a vocabulary disjoint from 
$\KB$ and $\phi$, then, for example, $\KB \dentailsn_\infp \Pr(\phi)
= \alpha$ iff $\KB
\land \KB' \dentailsn_\infp \Pr(\phi) = \alpha$.  If $\KB'$ is objective,
then the standard probabilistic approach would be to identify learning
$\KB'$ with conditioning
on $\KB'$.  Suppose that we restrict to
inference procedures that do indeed condition
on objective information (as is the case for the class of inference
procedures we consider in Section~\ref{limited}).
Then $\KB \land \KB' \dentailsn_\infp \Pr(\phi) =
\alpha$ exactly if $\KB \dentailsn_\infp \Pr(\phi \mid \KB') = \alpha$.
Thus, Theorem~\ref{noindep} tells us that inference procedures that
condition on new 
(objective)
information cannot both be representation independent
and ignore irrelevant information.

Thus, although representation independence, unlike robustness, does not
force us to use entirely trivial inference procedures, it does 
prevent us from using procedures 
that have certain highly desirable properties.

\section{Discussion}\label{discussion}
These results suggest that any type of representation independence is
hard to come by.  They also raise the concern that perhaps our
definitions were not quite right.  We can provide what seems to be even
more support for the latter point.  

\xam
Let $Q$ be a unary predicate and $c_1,\ldots,c_{100},d$ be constant
symbols.  Suppose that we have two vocabularies $\Phi = \{Q,d\}$ and
$\Psi = \{Q,c_1,\ldots,c_{100},d\}$.  Consider the interpretation
$i$ from $\Phi$ to $\Psi$ for which $i(d) = d$ and
$i(Q(x)) = Q(x) \land Q(c_1) \land \ldots \land
Q(c_{100})$.  
Now, consider $\KB = \exists x Q(x)$.  In this case, $i(\KB) = \exists x
(Q(x) \land Q(c_1) \land \ldots \land Q(c_{100})$.  Intuitively, since all
the $c_i$'s may refer to the same domain element, the only conclusion we can
make with certainty 
from
$Q(c_1) \land \ldots \land Q(c_{100})$ is that there
exists at least one $Q$ in the domain, which gives us no additional
information beyond $\KB$.  We can convert this example into a general
argument that the embedding $f$ corresponding to $i$ is faithful.
Intuitively, for any $\KB$, we can only get the conclusion $Q(c_1) \land
\ldots \land Q(c_{100})$ from $f^*(\KB)$ if $Q(x)$ appears
positively in $\KB$; but, in this case, we already know that there is at
least one $Q$, so we gain no new information from the embedding.
But it does not seem unreasonable that an inference procedure
should assign different degrees of belief to $Q(d)$ given  $\KB = \exists x
Q(x)$ 
on the one hand and given $i(\KB) = \exists x (Q(x) \land Q(c_1) \land
\ldots \land Q(c_{100}))$ on the other,%
\footnote{Actually, $i(Q(d)) = Q(d) \land Q(c_1) \land \ldots
\land
Q(c_{100})$,
but the latter is equivalent to $Q(d)$ given 
$\KB$.}
particularly if the domain is small.  In fact, many 
reasoning
systems explicitly adopt a {\em unique names assumption\/}, which would
clearly force different conclusions in these two situations.
\exam

This example suggests that, at least in the first-order case, even
faithful embeddings do not always match our
intuition for a ``reasonable'' representation shift.
One might therefore think that perhaps the problem is with our
definition even in the propositional case.  Maybe there is a totally
different definition of representation independence that avoids these
problems.  While this is possible, we do not believe it to be the case.
The techniques that we used to prove Theorem~\ref{noindep}
and~\ref{almosttrivial2} seem to apply to any reasonable notion of
representation independence.%
\footnote{They
certainly
applied to all of the many definitions that we tried!}
To give the flavor of the type of argument used to prove these
theorems, consider Example~\ref{colorful}, and
assume that $\true \dentailsn_\infp\, \Pr(\colorful) = \alpha$ for $\alpha \in
(0,1)$.%
\footnote{In fact, it suffices to assume that
$\true \dentailsn_\infp\, \Pr(\colorful) \in  [\alpha,\beta]$, as long as
$\alpha > 0$ or $\beta < 1$.}
Using an embedding $g$ such that $g(\colorful) = \red$, we
conclude that $\true \dentailsn_\infp\, \Pr(\red) = \alpha$.  Similarly, we can
conclude $\Pr(\blue) = \alpha$ and $\Pr(\green) = \alpha$.  But in order
for $\dentailsn_\infp$ to be invariant under our original embedding, we
must have $\true \dentailsn_\infp\, \Pr(\red \lor \blue \lor \green) =
\alpha$, which is completely inconsistent with our previous conclusions.
But the embeddings we use in this argument are very
natural ones; we would not {\em want\/} a definition of representation
independence that disallowed them.

These results can be viewed as support for the position that
representation dependence is justified; the choice of an appropriate
representation 
encodes significant information.
In particular, it encodes the bias of 
the knowledge-base designer about the world.  Researchers in machine
learning have long realized that bias is an inevitable component of
effective inductive reasoning
(i.e., learning from evidence).
So we should not be completely surprised if
it turns out that other types of leaping to conclusions (as in our context)
also depend on the bias.

But we need to be a little careful here.
For example, in some cases we can identify the vocabulary (and hence,
the representation) with the
sensors that an agent has at its disposal.  It may not seem that
unreasonable that an agent with a temperature sensor and a motion sensor
might carve up the world differently from an agent with a color sensor
and a distance sensor.  But consider two agents with different sensors
who have not yet made any observations.  Suppose that both of them can
talk about the distance to a tree.  Is it reasonable that the two agents
should reach different conclusions about the distance just because they
have different sensors (and thus use different vocabularies), although they
have not made any observations?  It would then follow that the agents
should change their conclusions if they switched sensors, despite not
having made any observations.  This does not seem so reasonable!

Bias and representation independence can be viewed as two extremes in
a spectrum.  If we accept that the knowledge base encodes the user's
bias, there is no obligation to be invariant under 
any representation shifts at all.  
On the other hand, if we assume that the
representation used carries no information, coherence requires that our
inference procedure give the same answers for all ``equivalent''
representations.
We believe that the right answer lies somewhere in
between.  There are typically a number of reasonable ways in which we can
represent our information, and we might want our inference procedure to
return the same conclusions no matter which of these we choose.
It thus makes
sense to require that our inference procedure be invariant under
embeddings that take us from one reasonable representation to another.  But
it does not follow that it must be invariant under {\em all\/}
embeddings, or even all embeddings that are syntactically similar to the
ones we wish to allow. We may be willing to refine $\colorful$ to $\red \lor
\blue \lor \green$ or to define $\flyingbird$ as $\fly \land \bird$, but not
to transform $\red$ to $\sparrow$.
\commentout{
More generally,
the symbols that we use typically
have linguistic
significance, and we often have some background knowledge about the
meaning of these symbols,
such as the fact that $\flyingbird$ is equivalent to $\fly \land\bird$.
This background knowledge should tell us which embeddings
are
``reasonable'' and which are not.  Our inference procedure should not be
forced to ``defend itself'' against embeddings that encode unreasonable
representation shifts.
Intuitively, by using background knowledge
to restrict the class of embeddings, we avoid
both of the problems that our definition of minimal representation
independence encountered:  By not requiring that the inference procedure
be invariant under {\em all\/} embeddings, we avoid being forced into
essential entailment.
Moreover, this ability to ``pick and choose'' the relevant embeddings
allows us to also require invariance under certain embeddings that are
not t-faithful.  In the next section, we show how we can use background
knowledge to construct inference procedures that are selectively
representation independent.
}
In the next section, we show how to construct inference procedures that are
representation independent under a limited class of representation shifts.

\section{Selective invariance}\label{limited}
As discussed above, we want to construct an inference procedure $\infp$
that is invariant only under certain embeddings.  For the purposes of this
section, 
we restrict attention to finite spaces, where all sets are measurable.
That is, we focus on $\X$-inference procedures where $\X$ consists only
of measure spaces of the form $(X,2^X)$, where $X$ is finite.

Our first step is to understand the conditions under which an $\X$-inference
procedure $\infp$ is invariant under a specific $X$-$Y$ embedding $f$.  When
do we conclude $\theta$ from $\KB \subseteq \Delta_X$?  Recall that an
inference procedure $\infp_X$ picks a subset $\D_X = \infp_X(\KB)$, and
concludes $\theta$ iff $\theta$ holds for every measure in
$\D_X$. Similarly, when applied to $f^*(\KB) \subseteq \Delta_Y$, $\infp_Y$
picks a subset $\D_Y = \infp_Y(f^*(\KB))$.  For $\infp$ to be invariant
under $f$ with respect to $\KB$, there has to be a tight connection between
$\D_X$ and $\D_Y$.

To understand this connection, first consider a pair of measures $\mu$
on $X$ and $\nu$ on $Y$.
Recall from Proposition~\ref{pro:correspond1} that $\mu$ and $\nu$
correspond under $f$ iff, for all formulas $\theta$, we have $\mu \sat
\theta$ iff $\nu \sat f^*(\theta)$.
To understand how the correspondence extends to sets of probability measures,
consider the following example:
\xam\label{correspondexample}
Consider the embedding $f$ of Example~\ref{embed.color}, and let $\D_X =
\{\mu,\mu'\}$
where $\mu$ is as above, and $\mu'(\colorful) = 0.6$.
How do we guarantee that we reach the corresponding
conclusions from $\D_X$ and $\D_Y$?  Assume, for example, that $\D_Y$
contains some measure $\nu$ that does not correspond to either
$\mu$ or $\mu'$,
e.g., the measure that assigns
probability $1/4$ to all four states.  In this case, the conclusion
$\Pr(\colorful) \le 0.7$ holds in $\D_X$, because it holds for both
these measures; but
the corresponding conclusion $\Pr(\red \lor \blue \lor \green) \le 0.7$
does not hold in $\D_Y$.  Therefore, every probability measure in $\D_Y$ must
correspond to some measure in $\D_X$.
Conversely, every measure in $\D_X$ must correspond to a
measure in $\D_Y$.  For suppose that there is no measure
$\nu \in \D_Y$ corresponding to $\mu$.  Then
we get the conclusion $\Pr(\blue \lor \red \lor \green)
\neq 0.7$ from $\D_Y$, but the corresponding conclusion
$\Pr(\colorful) \neq 0.7$ does not follow from $\D_X$.
Note that these two conditions do {\em not\/} imply that
$\D_Y$ must be precisely the set of measures corresponding to
measures in $\D_X$.  In particular, we might have $\D_Y$
containing only a single measure $\nu$ corresponding to $\mu$
(and at least one corresponding to $\mu'$), e.g., one with $\nu(\red)
= 0.5$, $\nu(\blue) = 0$, $\nu(\green) = 0.2$, and $\nu(\noncolorful)
= 0.3$.
\exam

Based on this example, we use the following extension to our definition of
correspondence.
\dfn
We say that $\D_X$ and $\D_Y$ {\em correspond\/} under $f$ if
for all $\nu \in \D_Y$, there exists a corresponding
$\mu \in \D_X$
(so that $\mu(S) = \nu(f(S))$ for all $S \subseteq X$),
and for all $\mu \in \D_X$,
there exists a corresponding 
$\nu \in \D_Y$.
\edfn
\pro\label{correspond2}
Suppose that $f$ is a faithful $X$-$Y$ embedding,  $\D_X
\subseteq \dists_X$, and $\D_Y \subseteq \dists_Y$.  The following two
conditions are equivalent: 
\begin{itemize}
\item[(a)] $\D_X$ and $\D_Y$ correspond under $f$;
\item[(b)] for all $\theta$, $\D_X \sat \theta$ iff $\D_Y \sat 
f^*(\theta)$.%
\footnote{While (a) implies (b) for arbitrary spaces, the implication
from (b) to (a) depends on the restriction to finite spaces made in this
section.  For suppose that $X$ is the natural numbers $\natnum$, $f$ is the
identity, $\D_X$ consists of all probability measures on $\natnum$, and
$\D_Y$ consists of all measures but that measure $\mu_0$ such that
$\mu_0(n) = 1/2^{n+1}$.  If the language consists of finite Boolean
combinations of assertions of the form $\Pr(S) \ge \alpha$, for $S
\subseteq \natnum$, then it is easy to see that $\D_X \sat \theta$ iff
$\D_Y \sat \theta$ for all formulas $\theta$, but clearly $\D_X$ and
$\D_Y$ do not correspond under the identity map.}
\end{itemize}
\epro
\prf See Appendix~\ref{sec6prfs}. \eprf

To produce an inference procedure that is invariant under some
$X$-$Y$ embedding $f$, we must ensure that {\em for every $\KB$}, 
$\infp_X(\KB)$ and $\infp_Y(\KB)$ correspond.  At first glance,
it seems rather difficult to guarantee correspondence for every knowledge
base.  It turns out that the situation is not that bad.  In the remainder of
this section, we show how, starting with a correspondence for the knowledge 
base $\true$---that is, starting with a correspondence between
$\infp_X(\Delta_X)$ and $\infp_Y(\Delta_Y)$---we can bootstrap to 
a correspondence for all $\KB$'s, using standard probabilistic updating
procedures.

First consider
the problem of
updating with objective information.  The standard way of doing this update
is via {\em conditioning}.  For a measure $\mu \in \dists_X$ and an
event $S \subseteq X$, define $\mu|S$ to be the measure that
assigns probability $\mu(w)/\mu(S)$ to every $w \in S$, and zero to all
other states.  For a set of measures $\D_X \subseteq \dists_X$, define
$\D_X|S$ to be $\{\mu|S\ :\ \mu \in \D_X\}$.  
The following result is easy to show.
\pro\label{conditioning}
Let $S \subseteq X$ be an event
and let $f$ be a faithful $X$-$Y$ embedding.
If $\mu$ and $\nu$ correspond under
$f$, then $\mu|S$ and $\nu|f(S)$ also correspond under $f$.
\epro
\prf Almost immediate from the definitions; left to the reader.  (In any
case, note that this result follows from Theorem~\ref{conditioning} below.)
\eprf

Clearly, the result extends to sets of measures.
\cor\label{cor:cross.entropy}
If $f$ is a faithful $X$-$Y$ embedding, and
$\D_X$ and $\D_Y$ correspond under $f$, then $\D_X|S$ and $\D_Y|f(S)$
also correspond under $f$. 
\ecor

What if we want to update on a constraint 
that is not objective?
The standard extension of conditioning to this case is via {\em
relative entropy\/} or {\em KL-divergence}~\cite{X.entropy}.
\dfn
If $\mu$ and $\mu'$ are measures on $X$, the {\em relative entropy
between $\mu'$ and $\mu$}, 
denoted $\KLD{X}{\mu'}{\mu}$, is defined as $\sum_{x \in X} \mu'(x)\log 
(\mu'(x) / \mu(x))$.  For a measure $\mu$ on $X$ and a constraint
$\theta$, let $\mu|\theta$ denote the set of measures
$\mu'$ satisfying $\theta$ for
which $\KLD{X}{\mu'}{\mu}$ is minimal.
\edfn
Intuitively, the KL-divergence measures the ``distance'' from $\mu'$ to
$\mu$.  A measure $\mu'$ satisfying $\theta$ for which
$\KLD{X}{\mu'}{\mu}$ is minimal can be thought of as the ``closest''
measure to $\mu$ that satisfies $\theta$.  If $\theta$ denotes an
objective constraint, then the unique measure satisfying $\theta$ for
which $\KLD{X}{\mu'}{\mu}$ is minimal is the conditional measure
$\mu|\theta$
\cite{X.entropy}.
(That is why we have deliberately used the same notation here
as for conditioning.)  
Moreover, it is easy to show that $\KLD{X}{\mu'}{\mu} = 0$ iff $\mu' =
\mu$.  It follows that if $\mu \in \theta$, then $\mu|\theta = \mu$.

Given a set of measure $\D_X \subseteq \dists_X$ and a constraint
$\theta$ on $\dists_X$, define $\D_X|\theta$ to be $\cup_{\mu \in \D_X}
\mu|\theta$.

We can now apply a well-known result (see, e.g., \cite{seidenfeld}) to
generalize Proposition~\ref{conditioning} to the case of relative entropy. 
\thm\label{relative.entropy}
Let $\theta$ be an arbitrary constraint on $\dists_X$.  If 
$f$ is a faithful $X$-$Y$ embedding and
$\mu$ and
$\nu$ correspond under $f$, then $\mu|\theta$ and $\nu|f^*(\theta)$ also
correspond under $f$.
\ethm

\prf See Appendix~\ref{sec6prfs}. \eprf

Again, this result clearly extends to sets of measures.
\cor\label{cor:relative.entropy}
If 
$f$ is a faithful $X$-$Y$ embedding, and
$\D_X$ and $\D_Y$ correspond under $f$, then $\D_X|\theta$ and
$\D_Y|f^*(\theta)$ also correspond under $f$.
\ecor

These results give us a way to ``bootstrap'' invariance.  We construct an
inference procedure that uses relative entropy starting from some set of
{\em prior probability measures}.  Intuitively, these encode the user's
prior beliefs about the domain.  As information comes in, these
measures are updated using cross-entropy.  If we design the priors so
that certain invariances hold, Corollary~\ref{cor:relative.entropy}
guarantees that these invariances continue to hold throughout the process.

Formally, a {\em prior function\/} $\P$ on $\X$ maps $X \in \X$ to a set
$\P(X)$ 
of probability measures in $\dists_X$.  
Define an inference procedure $\infp^\P$ by taking
$\infp^\P_X(\KB) =  
\P(X)|\KB$.
Note that $\infp^\P_X(\true) = \P(X)$, so that when we have
no constraints at all, we use $\P(X)$ as the basis for our inference. 
Most of the standard inference procedures are of the form $\infp^\P$
for some prior function $\P$.  It is fairly straightforward to verify, for
example, that entailment is 
$\infp^\P$
for $\P(X) = \dists_X$.
(This is because, as observed earlier, $\mu|\KB = \mu$ if $\mu \in \KB$.)
Standard Bayesian conditioning (defined for objective knowledge bases) is
of this form, where we take $\P(X)$ to be a single measure for
each space $X$.  More interestingly, it is well known~\cite{X.entropy} that
maximum entropy is $\infp^{\P_u}$ where $\P_u(X)$ is the singleton set
containing only the uniform prior on $X$.

So what can we say about the robustness of $\infp^\P$ to representation
shifts?  Using
Proposition~\ref{correspond2} and Corollary~\ref{cor:cross.entropy}, it
is easy to show 
that if we want $\I^\P$ to be
invariant under some set $\F$ of embeddings, then we must ensure that the
prior function has the right correspondence property.
\thm\label{bootstrap}
If $f$ is a faithful $X$-$Y$ embedding, then 
$\infp^\P$ is invariant under $f$
iff $\P(X)$ and $\P(Y)$ correspond under $f$.
\ethm
\prf See Appendix~\ref{sec6prfs}. \eprf

Theorem~\ref{bootstrap} sheds some light on the maximum entropy inference procedure.
As we mentioned, $\dentailssub{me}$ is precisely the inference
procedure based on the prior function $\P_u$.  The corollary asserts
that $\dentailssub{me}$ is invariant under $f$ precisely when the uniform
priors on $X$ and $Y$ correspond under $f$.  This shows that
maximum entropy's lack of representation independence
is an immediate
consequence of the identical problem for the uniform prior.  Is there a class
$\F$ of embeddings under which maximum entropy is invariant?
Clearly, the answer is yes.  It is easy to see that any embedding that takes
the elements of $X$ to (disjoint) sets of equal cardinality
has the correspondence property required by Theorem~\ref{bootstrap}.
It follows that maximum entropy is invariant under all
such embeddings.  In fact, the requirement that maximum entropy be
invariant under a subset of these embeddings is one of the axioms
in Shore and Johnson's \citeyear{ShoreJohnson}  axiomatic
characterization of maximum-entropy.
(We remark that Paris \citeyear[Theorem~7.10]{Paris94} proves that
maximum entropy satisfies a variant of his atomicity principle; his
invariance result is essentially a special case of Theorem~\ref{bootstrap}.)

If we do not like the behavior of maximum entropy under representation
shifts, Theorem~\ref{bootstrap} provides a solution:~we should simply start
out with a different prior function.  If we want to maintain invariance
under all representation shifts, 
$\P(X)$ must include all non-extreme priors (i.e., all the measures
$\mu$ on $X$ such that $\mu(A) \notin \{0,1\}$ for all $A$ such that $A
\notin \{\emptyset,X\}$).  This
set of priors gives essential entailment as an inference
procedure.  If, however, 
we have prior knowledge as to which embeddings encode ``reasonable''
representation shifts, we can often make do with a smaller class of priors,
resulting in an inference procedure that is more prone to leap to
conclusions.  Given a class of ``reasonable'' embeddings $\F$, we can often
find a prior function $\P$ 
that
is ``closed'' under each $f \in \F$, \ie
for each measure $\mu \in \P(X)$ and each $X$-$Y$ embedding $f \in
F$ we make sure that there is a corresponding measure $\nu \in \P(Y)$,
and vice versa.  Thus, we can guarantee that $\P$ has the appropriate
structure using a process of closing off under each $f$ in $\F$.

Of course, we can also execute this process in reverse. Suppose that we want to
support a certain reasoning pattern that requires leaping to
conclusions.  The classical example of such a reasoning pattern is, of
course, a default assumption of independence.  What is the ``most''
representation independence that we can get without losing this reasoning
pattern?  As we now show, Theorem~\ref{bootstrap} gives us the answer.

We begin by providing one plausible formulation of the desired 
reasoning pattern.  
For a finite space $X$, we say that $X_1 \times \cdots \times X_n$ is
the {\em product decomposition\/} of $X$ if $X = X_1 \times \cdots
\times X_n$ and $n$ is the largest number for which $X$ can be written
as a product in this way.  (It is easy to see that if $X$ is finite,
then this ``maximal'' product decomposition is unique.)
A measure $\mu \in \dists_X$ is a
\emph{product measure on $X$} if $X_1 \times \cdots \times X_n$ is the
product decomposition of $X$ and there
exist measures $\mu_i 
\in \dists_{X_i}$ for $i=1,\ldots,n$ such that 
$\mu = \mu_1 \times \cdots \times \mu_n$, that is, 
$\mu(U_1 \times \cdots \times U_n) = \prod_{i=1}^n \mu_i(U_i)$, if $U_i
\subseteq X_i$, $i = 1, \ldots, n$.
Let $\P_\Pi$ be the set of all product
measures on $X$.  If $\P_\Pi$ is the prior and the
relative entropy rule is used to update the prior given a knowledge
base, then $\dentailsn_{\P_\Pi}$ satisfies a form of minimal default
independence.  In fact, it is easy to show
that it satisfies the following stronger property. 
\pro\label{pro:prodmeasure}
Suppose that $X_1 \times \cdots \times X_n$ is the product decomposition
on $X$ and, for each $i=1,\ldots,n$, $\KB_i$ is  a constraint 
on $X_i$, and $S_i$ is a subset of $X_i$.  Then 
\[
\band_{i=1}^n \KB_i \dentailsn_{\infp_{\P_\Pi}} \Pr(S_1 \land \ldots \land
S_n) = \prod_{i=1}^n \Pr(S_i).
\]
\epro
\prf See Appendix~\ref{sec6prfs}. \eprf

Theorem~\ref{noindep} shows that $\dentailsn_{\P_\Pi}$ cannot be invariant
under all embeddings.  Theorem~\ref{bootstrap} tells us that it is
invariant under precisely those embeddings for which $\P_\Pi$ is invariant.
These embeddings can be characterized syntactically in a natural way.
Suppose that $\Phi_1, \ldots, \Phi_n$ is a partition of a finite set
$\Phi$ of primitive propositions.  Note that a truth assignment to the
primitive propositions in $\Phi$ can be  viewed as a ``crossproduct''
of truth assignments to the primitive propositions in $\Phi_1, \ldots,
\Phi_n$.  Under this identification, suppose that a set $X$ of truth
assignments to $\Phi$ is decomposed as $X_1 \times \cdots \times
X_n$, where $X_i$ consists of truth assignments to $\Phi_i$.  
In that case, if $p \in \Phi_j$ and $q, r \in \Phi_k$ for some $j \ne
k$, 
then 
$\true \dentailsn_\P \Pr(p \land q) = \Pr(p) \times \Pr(q)$, but since
since
$q$ and $r$ are in the same subset, we do not have 
$\true \dentailsn_\P \Pr(r \land q) = \Pr(r) \times \Pr(q)$.  
Hence, $\P_\Pi$ is not invariant under an interpretation $i$
that maps $p$ to $r$ 
and maps $q$ to itself.
Intuitively, the problem is that
$i$ is ``crossing subset boundaries''; it is mapping primitive
propositions that are in different subsets to the same subset.
If we restrict to interpretations 
that``preserve subset boundaries'', then we avoid this problem.

We can get a semantic characterization of this as follows.
If the product decomposition of $X$ is $X_1 \times \cdots \times X_n$
and the product decomposition of $Y$ is $Y_1 \times \cdots \times Y_n$, then
$f$ is an \emph{$X$-$Y$ product embedding} if $f$ is an $X$-$Y$
embedding and 
there are $X_i$-$Y_i$ embeddings $f_i$, $i = 1, \dots, n$, 
and $f(\langle x_1,\ldots,x_n \rangle) = f_1(x_1)
\times \cdots \times f_n(x_n)$.
Product embeddings capture the intuition of preserving subset
boundaries; elements in a given subset $X_i$ remain in the same subset
($Y_i$)
after the embedding.  However, the notion of product embedding is
somewhat restrictive; it requires that elements in the $i$th subset
of $X$ map to elements in the $i$th component of $Y$, for $i = 1, \ldots,
n$.  We can still preserve default independence if the components of a
product are permuted.
An $g$ is a {\em permutation embedding\/} if there exists a
permutation $\pi$ of $\{1, \ldots, n\}$ 
such that $g(\langle x_1,\ldots,x_n \rangle) = \langle x_{\pi(1)},
\ldots, x_{\pi(n)}\rangle$.

\thm\label{thm:products}
The inference procedure $\infp_{\P_\Pi}$ is invariant under faithful
product embeddings
and under permutation embeddings.
\ethm

\commentout{
An {\em independence structure\/} $\Pi$ over $\Phi^*$ (our
fixed infinite vocabulary) is a partition of the symbols in $\Phi^*$ into a
finite collection of disjoint sets or {\em cells\/}
$\Phi_1, \Phi_2, \ldots, \Phi_k$.  A measure $\mu$ on $\W(\Phi)$ {\em
respects the independence structure $\Pi$\/} if, for any formulas $\phi_i
\in \L(\Phi_i \inter \Phi)$ and $\phi_j \in\L(\Phi_j \inter \Phi)$
with $i \ne j$, we have $\mu(\phi_i \land \phi_j) = \mu(\phi_i)\mu(\phi_j)$.
Thus, $\mu$ makes the denotations of the symbols in
different cells independent.  Let $\P^\Pi(\Phi)$ be the class of all
measures $\mu$ on $\W(\Phi)$ that respect $\Pi$.
We can prove that $\dentailsn_{\P_\Pi}$
enforces minimal default independence for symbols in different cells.
In fact, it satisfies the following stronger property:
\pro
Let $\Psi_1$ and $\Psi_2$ be disjoint vocabularies each of which is
the union of cells in $\Pi$.
If $\KB_1 \in \LPr(\Psi_1)$, $\theta_1 \in \L(\Psi_1)$,
$\KB_2 \in \LPr(\Psi_2)$, $\theta_2 \in \L(\Psi_2)$, then
$\KB_1 \land \KB_2 \dentailsn_{\P_\Pi}
\Pr(\theta_1 \land \theta_2) = \Pr(\theta_1) \times \Pr(\theta_2)$.
\epro
This result follows directly from the Shore and
Johnson~\citeyear{ShoreJohnson} axiomatization of relative entropy.

Theorem~\ref{noindep} shows that $\dentailsn_{\P_\Pi}$ cannot be invariant
under all embeddings.  Theorem~\ref{bootstrap} tells us that it is
invariant under precisely those embeddings for which $\P_\Pi$ is invariant.

Since $p$ and $q$ are in different cells, we have
$\true \dentailsn_\P \Pr(p \land q) = \Pr(p) \times \Pr(q)$.
However, since
$q$ and $r$ are in the same cell, we do not have 
$\true \dentailsn_\P \Pr(r \land q) = \Pr(r) \times \Pr(q)$.  
Hence, $\P_\Pi$ is not invariant under an interpretation $i$
that maps $p$ to $r$.  Intuitively, the problem is that
$i$ is ``crossing cell boundaries''.  If we restrict to interpretations $i$
that do not cross cell boundaries, i.e., those that for any $p \in \Phi_i$
have $i(p) \in \L(\Phi_i)$, then we avoid this problem.
\thm
The inference procedure $\dentailsn_{\P_\Pi}$ is invariant under
any interpretation $i$ that is faithful and does not cross cell boundaries.
\ethm
}%

Theorem~\ref{bootstrap} thus provides us with the basic tools to easily
define an inference procedure that enforces minimal default independence
for constraints involving disjoint parts of the language, while at the same
time guaranteeing invariance under a large and natural class of
embeddings.   Given our negative result in Theorem~\ref{noindep}, this type
of result is the best that we could possibly hope for.  In general,
Theorem~\ref{bootstrap} provides us with a principled framework for
controlling the tradeoff between the 
strength of the conclusions that can be reached by an inference
procedure and invariance under representation shifts.

\section{Related Work}\label{relatedwork}
As we mentioned earlier, there are two types of probabilistic inference.
We partition our discussion of related work along those lines.

\subsection{Probabilistic Inference from a Knowledge Base}
Given the importance of representation in reasoning, 
and the fact that one of the
main criticisms of maximum entropy has been its sensitivity to
representation shifts, it is surprising how little work there
has been on the problem of representation dependence.
Indeed, to the best of our knowledge, the only work 
that has focused 
on representation
independence in the logical sense that we have considered here 
prior to ours
is that
of Salmon
and Paris.

Salmon \citeyear{Salmon1} defined a {\em criterion of linguistic
invariance}, which seems essentially equivalent to our notion of
representation independence.
He tried to use this criterion to defend one particular method of
inductive inference but, as pointed out by Barker in the commentary
at the end of~\cite{Salmon1}, his preferred method does not
satisfy his criterion either.  Salmon \citeyear{Salmon2}
then attempted to define
a modified inductive inference method that would satisfy his criterion
but it is not clear that this attempt succeeded.
In any case, our results show that his
modified method certainly cannot be representation independent in our
sense.

As we said earlier, Paris \citeyear{Paris94} considers 
inference processes, which given a constraint on $\dists_X$, choose a  
unique measure satisfying the constraint.  He then considers various
properties that an inference process might have.  Several of these are
closely related to properties that we have considered here.  (In
describing these notions, we have made some inessential changes so as to
be able to express them in our notation.)
\begin{itemize}
\item An $\X$-inference process $I$ is {\em language
invariant\/} if 
all $X, Y \in \X$ and all
constraints $\KB$ and $\phi$ on $\dists_X$, we have that 
$\KB \dentailsn_{I_X} \phi$ iff $\KB \dentailsn_{I_{X \times
Y\,}} \phi$.  Clearly language invariance is a special case of
robustness.  Paris shows that a {\em center of mass\/}
inference process (that, given a set $A \subseteq \dists_X$, chooses the
measure that is the center of mass of $A$) is not language invariant;
on the other hand, it is well known that maximum entropy is language
invariant.  
\item An $\X$-inference process $I$ satisfies the {\em principle of
irrelevant information\/} if for all spaces $X, Y \in \X$,  
constraints $\KB$ and $\phi$ on $\dists_X$, and constraints $\psi$
on $\dists_{Y}$, we have
$\KB \dentailsn_{I_X} \phi$ iff $\KB \land \psi \dentailsn_{I_{X \times
Y\,}} \phi$.  Again, this is a special case of robustness, since a
constraint $\psi$ on $\dists_{Y}$ must be $X$-conservative.  Paris shows
that maximum entropy satisfies this principle.  (He restricts the domain
of the maximum entropy process to closed convex sets, so that there is
always a unique probability measure that maximizes entropy.)
\item An $\X$-inference process $I$ satisfies the {\em renaming
principle\/} if, whenever $X$ and $Y$ are finite spaces, $g: X
\rightarrow Y$ is an isomorphism, and $f: 2^X \rightarrow 2^Y$ is the
faithful embedding based on $g$ (in that $f(S)  = \{g(s): s \in S\}$),
then for all constraints 
$\KB$ and $\theta$ on $\dists_X$, we have $\KB \dentailsn_{\infp_X\,}
\theta$ iff $f^*(\KB) 
\dentailsn_{\infp_Y}\, f^*(\theta)$.  Clearly, the renaming principle is
a special case of representation independence.  Paris shows that a
number of inference processes (including maximum entropy) satisfy the
renaming principle.
\item An $\X$-inference process $I$ satisfies the {\em principle of
independence\/} if, whenever $X$, $Y$, and $Z$ are in $\X$, $S \in
\F_X$, $T \in \F_Y$, $U \in \F_Z$, and $\KB$ is the constraint
$\Pr(U) = a \land \Pr(S | U) = b \land \Pr(T | U) = c$, where $a > 0$,
then $\KB \dentails \Pr(S \times T | U) = bc$.  Ignoring the conditional
probabilities, this is clearly a special case of minimal default
independence.  Paris and Vencovska \citeyear{ParisV90} show that maximum
entropy is the unique inference process satisfying a number of
principles, including renaming, irrelevant information, and
independence.
\item An $\X$-inference process $I$ satisfies the {\em atomicity
principle} if, for all $X$, $Y_1$, \ldots, $Y_n$ in $\X$,
whenever $f'$ is an embedding from $\{0,1\}$ to 
$X$, and $f$ is the obvious extension of $f'$ to an embedding from
to $\{0,1\} \times Y_1 \times  \ldots \times Y_n$ to $X \times Y_1
\times \ldots \times Y_n$, then for all constraints $\KB$ and $\theta$
on $\dists_{\{0,1\} \times Y_1 \times \ldots \times Y_n}$, we have $\KB
\dentailsn_{\infp_X\,}  \theta$ iff $f^*(\KB) \dentailsn_{\infp_Y}\,
f^*(\theta)$.  Clearly atomicity is a special case of representation
independence.  Paris shows that there is no inference process that
satisfies atomicity.  The argument is similar in spirit to that used 
to prove Theorems~\ref{almosttrivial1} and~\ref{noindep}, but much
simpler, since inference 
processes return a unique probability measure, not a set of them.
\end{itemize}

More recently, Jaeger  \citeyear{Jaeger96}, building on our definitions,
has examined
representation independence for general nonmonotonic logics.  He
considers representation independence with respect to a collection of
transformations, and proves results about the degree to which certain
nonmonotonic formalisms, such as {\em rational closure}
\cite{LehmannMagidor}, satisfy representation independence.

Another line of research that is relevant to representation independence
is the work on {\em abstraction\/} \cite{GW92,NL94}.
Although the goal of this work is again to make connections between two
different ways of representing the same situation, there are significant
differences in focus.  In the work on abstraction, the two ways of
representing the situation are not expected to be equivalent.
Rather, one representation typically abstracts away irrelevant details that
are present in the other.  On the other hand, their treatment of the issues
is in terms of deductive entailment, not in terms of general inference
procedures.  It would be interesting to combine these two lines of work.

\subsection{Bayesian Probabilistic Inference}\label{sec:Bayesianapproach}
Bayesian statistics takes a very different perspective on the issues we
discuss in this paper.  As we discussed, the Bayesian approach generally
assumes that we construct a prior, and use standard probabilistic
conditioning to update that prior as new information is obtained.  In this 
approach, the representation of the knowledge obtained has no effect on the
conclusions.  Two pieces of information that are semantically equivalent
(denote the same event) will have precisely the same effect when used to
condition a distribution.  

In this paradigm, our analysis is more directly related to the step that  
precedes the probabilistic conditioning---the selection of the prior.
When we have very specific beliefs that we want to encode in a
prior distribution (as we do, for example, when constructing a Bayesian
network), we design our prior to reflect these beliefs in terms of the
vocabulary used.  For example, if we have a particular distribution in mind
over the location of an object, we will encode it one way when representing
the space in terms of Cartesian coordinates, and in another way when
using polar coordinates.  In effect, we can view the representation
transformation as an embedding $f$, and the two priors as corresponding
under $f$, in the sense of Definition~\ref{dfn:correspond}.  Thus, the
design of the prior already takes the representation into account.

On the other hand, when we are trying to construct an ``uninformed'' prior
for some class of problems, the issue of representation
independence becomes directly relevant.  Indeed, most of the standard
problems with maximum entropy arise even in the simple case
when we simply do Bayesian conditioning starting with a uniform prior over
our space.

A standard approach in Bayesian statistics is to use the invariance under
certain transformations in order to define an appropriate uninformed
prior.  For example, we might want a prior over images that is invariant to
rotation and translation.  In certain cases, once we specify the
transformation under which we want a measure to be invariant, the measure
is uniquely determined \cite{Jaynes68,KassWasserman}.  In this case, the
argument goes, the uniquely determined measure is perforce the ``right''
one.  This idea of picking a prior using its invariance properties is in
the same spirit as the approach we take in Section~\ref{limited}.  Indeed,
as this approach simply uses standard probabilistic conditioning for
objective information (such as observations), the Bayesian approach with an
uninformed prior invariant to a set of embeddings is, in a sense, a special
case.  However, our approach does not force us to choose a unique prior.
Rather, we allow the use of a set of prior
distributions, allowing us to explore a wider spectrum of inference
procedures.  

This approach is also related to the work of Walley \citeyear{Walley96},
who observes that representation independence is an important desideratum
in certain statistical applications involving multinomial data.  Walley
proposes the use of sets of Dirichlet densities to encode ignorance about a
prior, and shows that this approach is representation independent in its
domain of application.

\commentout{
Although statisticians have not considered representation independence
in the sense we have defined it here, Bayesian statisticians
have been very concerned with 
the
related issue of invariance
under certain transformations of parameters.  For example, we would expect
that our beliefs about a
person's height should be invariant under a transformation from feet to
meters.  
As we observed, such transformations can be understood as embeddings.
Since Bayesian statisticians are typically interested in choosing a
unique probability measure as a prior, 
their hope is that once we specify the transformation under
which we want a measure to be invariant, the measure will be
uniquely determined \cite{Jaynes68,KassWasserman}.  In this case, the
argument goes, the uniquely determined measure is perforce the
``right'' one.  This idea of picking a measure using its invariance
properties is in the same spirit as the approach we take in
Section~\ref{limited}.  But unlike the standard Bayesian approach, 
our approach does not force a choice of a unique measure.
This flexibility
enables us to explore a wider spectrum of inference procedures.

Walley \citeyear{Walley96} does not look at the problem of
representation independence in general, but observes that it is an
important desideratum in certain statistical applications involving
multinomial data.  He then
defines a particular inference procedure involving sets of Dirichlet
measures, and shows that it is representation independent in its domain
of application.

}

\section{Conclusions}\label{conclusions}
This paper takes a first step towards understanding the issue of
representation dependence in probabilistic reasoning,
by defining notions of invariance and representation
independence, showing that representation independence
is incompatible with drawing many standard default conclusions, and defining
limited notions of invariance that might that allow a compromise
between the desiderata of being able to draw interesting conclusions 
(not already entailed by the evidence) 
and representation independence.
Our focus here has been on inference in probabilistic logic, but
the notion of representation independence is just as important in
many other contexts.  Our definitions can clearly be extended to
non-probabilistic logics.  
As we mentioned, Jaeger \citeyear{Jaeger96} has obtained some results 
on representation independence in a more general setting, but there is
clearly much more that can be done.  
More generally, it would be of interest to understand better the tension
between representation independence and the strength of conclusions
that can be drawn from an inference procedure.

\acks{
Thanks to Ed Perkins for pointing us to \cite{KT64} and, in
particular, the result that a 
countably additive probability measure defined on a subalgebra of an
algebra $\F$ could not necessarily be extended to a countably additive
probability measure on $\F$.
Thanks to the reviewers of the paper for their perceptive comments and
for pointing out \cite{HT48}.
Much of Halpern's work on the paper was done while he was at the IBM
Almaden Research Center.  His recent work has been supported by NSF
under grant IRI-96-25901 and IIS-0090145 and ONR under grant N00014-01-1-0795.
Some of Koller's work was done at
U.C.\ Berkeley.  
Her research was sponsored in part by the Air Force Office of
Scientific Research (AFSC), under Contract F49620-91-C-0080, and by a
University of California President's Postdoctoral Fellowship.
Daphne Koller's 
later work on the paper
was supported through the generosity
of the Powell foundation, and by ONR grant N00014-96-1-0718.
A preliminary version of this appears in {\em Proceedings of
IJCAI~'95}, pp.~1853--1860.
}

\appendix

\section{Proofs}

\subsection{Proofs for Section~\protect{\ref{procedures}}}\label{sec2prfs}

\opro{choosedists.side2}
Let $\dentails$ be a relation on probabilistic constraints 
on $X$ for
which the properties Reflexivity, Left Logical Equivalence,
Right Weakening, Infinitary And, and Consistency hold
for all $\KB$ in the domain of $\dentails$.  (That is, if $\KB$ is in
the domain of $\dentails$, in that $\KB \dentails \theta$ for some
$\theta$, then $\KB \dentails \KB$, and so on.)
Then $\dentails$ is $\dentailsn_{\infp}$ for some $X$-inference procedure 
$\infp$.
\eopro
\prf Define $I$ as follows.  If $A \subseteq \dists_X$,
$\KB$ is in the domain of $\dentails$,
and $A = \intension{\KB}{X}$ for some statement $\KB$, then 
$A$ is in the domain of $I$ and
$I(A) =
\inter\{\intension{\theta}{X}: \KB \dentails \theta\}$.  Note that by
Left Logical Equivalence, this is well defined, since if $A =
\intension{\KB'}{X}$, then $\inter\{\intension{\theta}{X}: \KB \dentails
\theta\} = \inter\{\intension{\theta}{X}: \KB' \dentails \theta\}$.
If $A \ne \intension{\KB}{X}$ for some statement $\KB$, then 
$A$ is not in the domain of $I$.
It remains to check that $I$ is an $X$-inference
procedure (i.e., that $I(A) \subseteq A$ and that $I(A) = \emptyset$
iff $A = \emptyset$
for all $A$ in the domain of $I$),
and that $\dentails = \dentailsn_{I}$.  To check
that $I$ is an $X$-inference procedure, 
suppose that $A$ is in the domain of $I$.  Thus, $A = \intension{\KB}{X}$
By Reflexivity, it
easily follows 
that $I(\intension{\KB}{X}) \subseteq \intension{\KB}{X}$.  Next suppose that
$I(\intension{\KB}{X}) = \emptyset$.
It follows that $\inter\{\intension{\theta}{X}: \KB \dentails \theta\} =
\emptyset$.  Thus, $\{\theta: \KB \dentails \theta\} \sat \false$.  By
the Infinitary AND rule, we must have $\KB \dentailsn_{\infp}\, \false$.  
By the Consistency Rule, it follows that $\intension{\KB}{X} =
\emptyset$.  Thus, $I$ is indeed an $X$-inference procedure.  Finally,
note that if $\KB \dentailsn\, \psi$ then, by definition of $I$,
$I(\intension{\KB}{X}) \subseteq \intension{\psi}{X}$, so 
$\KB \dentailsn_{I}\, \psi$.  For the opposite inclusion, note that
if $\KB \dentailsn_{I}\, \psi$, then $\{\theta: \KB \dentails \theta\}
\sat \psi$.  Thus, by the Infinitary And rule, it follows that $\KB
\dentails_{\infp\,} \psi$.  \eprf

\subsection{Proofs for Section~\protect{\ref{sec:robust}}}\label{sec3prfs}

To prove Theorem~\ref{almosttrivial2}, we need the following lemma.

\lem\label{useful} Given two 
spaces $X_0$ and $X_1$, measures $\mu^0 \in \dists_{(X_0,\F_{X_0})}$ and
$\mu^1 \in \dists_{(X_1,\F_{X_1})}$, and subsets $S_0 \in \F_{X_0}$ and $S_1
\in \F_{X_1}$ such that 
$\mu^0(S_0) = \mu^1(S_1)$, there exists a measure
$\mu^2 \in \dists_{(X_0 \times X_1, \F_{X_0 \times X_1})}$ such that
$\mu^2_{X_i} = \mu^i$, for 
$i = 1,2$, and $\mu^2(S_0 \dimp S_1) = 1$.%
\footnote{If $A$ and $B$ are sets, we use the notation $A \dimp B$ to
denote the set $(A \inter B) \union (\overline{A} \inter \overline{B})$.}
\elem

\prf
\commentout{
For $(x_0,x_1) \in X_0 \times X_1$, we define
$$\mu^2((x_0,x_1)) =
\left \{\begin{array}{ll}
\mu^0(x_0)\mu^1(x_1)/\mu^1(S_1)) &\mbox{if $x_0 \in S_0$, $x_1
\in S_1$, and $\mu^1(S_1) \ne 0$}\\
\mu^0(x_0)\mu^1(x_1)/\mu^1(\overline{S_1}) &\mbox{if $x_0 \notin
S_0$, $x_1 \notin S_1$, and $\mu^1(\overline{S_1}) \ne 0$}\\
0 &\mbox{otherwise.} \end{array} \right .
$$
Clearly $$\begin{array}{ll}
&\sum_{(x_0,x_1) \in X_0 \times X_1} \mu^2((x_0,x_1))\\
= &\sum_{x_0 \in S_0} \sum_{x_1 \in X_1} \mu^2((x_0,x_1)) +
\sum_{x_0 \in \overline{S_0}} \sum_{x_1 \in X_1} \mu^2((x_0,x_1))\\
= &\sum_{x_0 \in S_0} \sum_{x_1 \in S_1} \mu^0(x_0)\mu^1(x_1)/\mu^1(S_1)
+ \sum_{x_0 \in \overline{S_0}} \sum_{x_1 \in \overline{S_1}}
\mu^0(x_0)\mu^1(x_1)/\mu^1(\overline{S_1})\\
= &\sum_{x_0 \in S_0} \mu^0(x_0)\sum_{x_1 \in S_1} \mu^1(x_1)/\mu^1(S_1)
+ \sum_{x_0 \in \overline{S_0}} \mu^0(x_0)\sum_{x_1 \in \overline{S_1}}
\mu^1(x_1)/\mu^1(\overline{S_1})\\
= &\sum_{x_0 \in S_0} \mu^0(x_0) +
\sum_{x_0 \in \overline{S_0}} \mu^0(x_0)\\
= &\mu^0(S_0) + \mu^0(\overline{S_0})\\
= &1.
\end{array}$$
A similar (and simpler) argument works if $\mu^1(S_1) = 0$
or $\mu^1(\overline{S_1}) = 0$ (for then $\mu^0(S_0) = 0$ or
$\mu^0(\overline{S_0}) = 1$, by assumption).

It is also clear that
$\mu^2(S_0 \dimp S_1) = 1$, since the only pairs to which
$\mu^2$ assigns positive probability are those in $(S_0 \times
S_1) \union (\overline{S_0} \times \overline{S_1})$.
Note that for $x_0 \in S_0$, we have
$\mu^2(\{x_0\} \times X_1) = \mu^0(x_0) \mu^1(S_1)/\mu^1(S_1) =
\mu^0(x_0)$, while for
$x_0 \in \overline{S_0}$, we have
$\mu^2(\{x_0\} \times X_1) = \mu^0(x_0)
\mu^1(\overline{S_1})/\mu^1(\overline{S_1}) = \mu^0(x_0)$.  Thus,
$\mu^2_{X_0} = \mu^0$.  Finally,
note that if $x_1 \in S_1$, then $\mu^2(X_0\times \{x_1\}) =
\mu^0(S_0) \times \mu^1(x_1)/\mu^1(S_1)$ while
if $x_1 \in \overline{S_1}$, then $\mu^2(X_0\times \{x_1\}) =
 \mu^0(\overline{S_0}) \times \mu^1(x_1)/\mu^1(\overline{S_1})$.
Since, by assumption, $\mu^0(S_0) = \mu^1(S_1)$, we get that
$\mu^2(X_0 \times \{x_1\}) = \mu^1(x_1)$, so $\mu^2_{X_1} = \mu^1$.
\eprf}%
For $A \times B \in \F_{X_0} \times \F_{X_1}$, define
$$\mu^2(A \times B) = (\mu^0(A \inter S_0) \mu^1(B \inter
S_1)/\mu^1(S_1)) + (\mu^0(A \inter \overline{S_0}) \mu^1(B \inter
\overline{S_1})/\mu^1(\overline{S_1})),$$ where we take 
$\mu^0(A \inter S_0) \mu^1(B \inter S_1)/\mu^1(S_1) = 0$ if $\mu^1(S_1) =
0$ and take 
$\mu^0(A \inter \overline{S_0}) \mu^1(B \inter
\overline{S_1})/\mu^1(\overline{S_1}) = 0$ if $\mu^1(\overline{S_1}) = 0$.
Extend to disjoint unions of such sets by additivity.  
Since all sets in $\F_{X_0 \times X_1}$ can be written as disjoint unions of
sets of the form $A \times B \in \F_{X_0} \times \F_{X_1}$,
this suffices to define $\mu^2$.
To see that $\mu^2$ is actually a measure, note that $\mu^2(X \times Y)
= \mu^0(S_0) + \mu^0(\overline{S_0}) = 1$.  Additivity is clearly
enforced by the definition.  Finally, to see that $\mu^2$ has the
desired properties, suppose that $\mu^1(S_1) \ne 0$ and
$\mu^1(\overline{S_1}) \ne 0$.  (The argument is easier if this is not
the case; we leave details to the reader.)  Then
$$\begin{array}{ll}
\mu^2_{X_0}(A) = \mu^2(A \times Y) =
\mu^0(A \inter S_0)\mu^1(S_1)/\mu^1(S_1) + \mu^0(A \inter \overline{S_0}
\mu^1(\overline{S_1})/\mu^1(\overline{S_1}) \\
\ \ \ = \mu^0(A \inter S_0) +
\mu^0(A \inter \overline{S_0}) = \mu^0(A).
\end{array}
$$
Since $\mu^0(S_0) = \mu^1(S_1)$ by assumption (and so $\mu^0(\overline{S_0}) =
\mu^1(\overline{S_1})$), 
$$\begin{array}{ll}
\mu^2_{X_1}(B) = \mu^2(X \times B) =
\mu^0(S_0) \mu^1( B \inter S_1)/\mu^1(S_1) + \mu^0(\overline{S_0})
\mu^1(B \inter \overline{S_1})/\mu^1\overline{S_1}) \\
\ \ \ = \mu^1( B \inter S_1) + \mu^1(B \inter \overline{S_1}) = \mu^1(B).
\end{array}$$

This completes the proof.
\eprf

\medskip

\othm{almosttrivial2}
If $\{I_X: X \in \X\}$ is a robust $\X$-inference procedure
that satisfies DI1, DI2, and DI3,
then $I_X$
is essentially entailment for all $X \in \X$.
\eothm

\prf  Suppose that $\{I_X: X \in \X\}$ is robust and $I_X$
is not essentially entailment for $X \in \X$. 
Then there must be a 
constraint $\KB$ on $\dists_X$ and a set $S \in \F_X$ such
that $\KB \dentailsn_{I\,} \alpha < \Pr(S) < \beta$ and
$\KB \not\sat \alpha \le \Pr(S) \le \beta$.  Thus, there must be
some $\gamma \notin [\alpha, \beta]$ such that $\KB \land \Pr(S) =
\gamma$ is consistent.  We can assume without loss of generality that
$\gamma < \alpha$ (otherwise we can replace $S$ by $\overline{S}$).

We first construct a space $Y_0 \in \X$ that has subsets
$U_1, \ldots, U_n$ with the following properties:
\begin{enumerate}
\item[(a)] There is no measure $\mu \in \dists_{Y_0}$ such that
$\mu(U_i) > \alpha$, for 
all
$i = 1, ..., n$.
\item[(b)] For each $i$, there is some measure $\mu_i' \in \dists_{Y_0}$
such that
$\mu_i'(U_i) = 1$ and $\mu_i'(U_j) > \gamma$ for all $j \ne i$.
\end{enumerate}
We proceed as follows.
Choose $n$ and $d$ such that $\gamma < (d-1)/(n-1)
< d/n < \alpha$.  By assumption, there exists a $Y_0 \in \X$ 
such that $|Y_0| =  n!/(n-d)!$.  Without loss of
generality, we can assume that $Y_0$ consists of all
tuples of the form $(a_1, \ldots, a_d)$, where the $a_i$'s are all
distinct, and between 1 and
$n$.  Let $U_i$ be 
consist
of all the tuples in $Y_0$ that have
$i$ somewhere in the subscript; it is easy to see that there are
$d(n-1)!/(n-d)!$ such tuples.
Suppose that $\mu$ is a probability measure in $\dists_{Y_0}$.
It is easy to see that $\mu(U_1) + \cdots + \mu(U_n) = d$, since
each tuple in $Y_0$ 
is in exactly $d$ of the $U_i$'s and so 
gets counted exactly $d$ times, and the sum of the
probabilities of the tuples is 1.  Thus, we cannot have
$\mu(U_i) > d/n$ for all $i$ (and, {\em a fortiori}, we cannot have
$\mu(U_i) > \alpha$ for all $i$).
This takes care of the first requirement.  Next, consider a
probability distribution $\mu_i'$ that
makes all the tuples making up $U_i$ equally probable, and gives all
the other tuples probability 0.  Then it is easy to see that
$\mu_i'(U_i) = 1$.  Moreover, since it is straightforward to
check that there are exactly $d(d-1)(n-2)!/(n-d)!$ tuples in $U_i \inter
U_j$ for $j \ne i$, we have $\mu_i'(U_j)
= [d(d-1)(n-2)!/(n-d)!]/[d(n-1)!/(n-d)!] = (d-1)/(n-1)$.  This takes care of
the second requirement.

By assumption, there is also a measurable space $Y \in \X$ such
that $|Y| = 2$.  Suppose that $Y = \{y,y'\}$.  Let
$Z = X^n \times Y_0 \times Y^n$, where 
the $n$ is the same as the $n$ chosen in the construction of $Y_0$.
Again, by assumption, $Z \in \X$.
For $i = 1, \ldots, n$, 
\begin{itemize}
\item if $A \subseteq X$, let 
 $A_i = X^{i-1} \times A \times X^{n-i} \times Y_0 \times Y^n \subseteq Z$. 
\item let $\KB_i = \{\mu \in \dists_Z: \mu_{X_i} \in \KB\}$;
\item let $Y_i$ be the subset of $Y^n$ where the $i$th copy of
$Y$ is replaced by $\{y\}$;  
\item  let $V_i$ be the subset of $Z$
of the form $X^n \times U_i \times Y_i$ (where $U_1, \ldots, U_n$ are
the subsets of $Y_i$ constructed above).
\end{itemize}
Let $\sigma$ be the following constraint on $\dists_Z$:
$$\KB_1 \land \ldots \land \KB_n \land 
\Pr(S_1 \dimp V_1) = 1 \land \ldots \land \Pr(S_n \dimp V_n) = 1.$$

Let $X_i$ denote the $i$th copy of $X$ in $Z$.  That is, for ease of
exposition, we view $Z$ as being of the form $X_1 \times \cdots \times
X_n \times Y_0 \times Y^m$, although all the $X_i$'s are identical,
since it is helpful to be able to refer to a specific $X_i$.
We claim that $\sigma$ is
$X_i$-conservative over $\KB$, for $i = 1, \ldots, n$.
Thus, we must show that
$\proj_{X_i}(\intension{\KB_i \land \sigma}{Z})
= \intension{\KB}{X}$.
It is immediate that
$\proj_{X_i}(\intension{\KB_i \land \sigma}{Z})
\subseteq \intension{\KB}{X}$.
For the
opposite inclusion, suppose that $\nu \in \intension{\KB}{X}$.
We must show that there exists some $\mu \in \intension{\KB_i \land
\sigma}{Z}$ such that $\mu_{X_i} =
\nu$. We proceed as follows.

Let $\mu'_0$ be a measure in $\dists_{Y_0}$ such that
$\mu'_0(U_i) = 1$ and $\mu'_0(U_j) > \gamma$, for $j \ne i$.
By construction of the $U_j$'s, such a measure must exist.
For $j \in \{1, \ldots, n\}$, let $\mu'_j$ be the measure in
$\dists_{Y}$ such that $\mu'_i(y) = \nu(S)$ and 
if $j \ne i$, then $\mu'_j(y) =
\gamma/\mu'_0(U_j)$ (and $\mu'_j(y') = 1 - \mu'_j(y)$).   Let $\mu'$ 
be the measure
on $Y_0 \times Y^n$ that is the ``crossproduct'' of $\mu'_0, \mu'_1,
\ldots, \mu'_n$. 
That is, $\mu'(T_0 \times \cdots \times T_n) = \mu'_0(T_0) \times \cdots
\times \mu'_n(T_n)$.  By construction, 
$\mu'(V_j) = \gamma$ for $j \ne i$ and $\mu'(V_i) = \nu(S)$.

By assumption, there is a measure $\nu_0 \in \dists_X$ such that
$\nu_0 \sat \KB \land \Pr(S) = \gamma$.
We now proceed inductively to define
a measure $\mu^k \in \dists_{X^k \times Y_0 \times Y^n}$ such that
(a) $\Pr((S_1 \dimp V_1) \inter 
\ldots \inter (S_k \dimp V_k)) = 1$, 
(b) $\mu^j_{Y} = \mu'$ and $\mu^{j}_{X_j} = \nu$ for $j = 1,
\ldots, k$.  We define $\mu^0 = \mu'$.
For the inductive step, we simply apply Lemma~\ref{useful}.
Finally, we take $\mu$ to be $\mu^n$.
Our construction guarantees that $\mu_{X^j} = \nu$, hence that
$\mu \sat \KB_j$.
In addition, the construction
guarantees that $\mu \sat
\Pr(S_1 \dimp V_1) = 1 \land \ldots \land  \Pr(S_n \dimp V_n) = 1$.
Hence $\mu \sat \sigma$, as desired.

It follows from DI1, DI2, and DI3 that $\sigma$ is in the domain of $I_{Z}$. 
Since $\KB_i \land \sigma$ is equivalent to $\sigma$, it follows that $\KB_i
\land \sigma$ is also in the domain of $I_Z$.
Now, by robustness, for any constraint $\phi$ on
$\dists_{X_i}$, we have $\KB_i \land \sigma \dentailsn_{I\,} \phi$ iff $\KB_i
\dentailsn_{I\,} \phi$.  Since $\KB_i \dentailsn_{I\,} \Pr(S_i) > \alpha$ and
$\KB_i \land \sigma$ is equivalent to $\sigma$, it follows that
$\sigma \dentailsn_{I\,} \Pr(S_i) > \alpha$ for $i = 1, \ldots, n$.
By the And rule (Proposition~\ref{choosedists.side1}), it follows that
$\sigma \dentailsn_{I\,} \Pr(S_1) > \alpha \land \ldots \land \Pr(S_n) >
\alpha$.
Since $\sigma \sat \Pr((S_1 \dimp V_1) \inter
(S_n \dimp V_n) ) = 1$,
it easily follows that
$\sigma \dentailsn_{I\,} \Pr(U_1) > \alpha \land \ldots \land \Pr(U_n) >
\alpha$.  But
our construction guarantees that $\Pr(U_1) > \alpha \land \ldots
\land \Pr(U_n) > \alpha$ is inconsistent.  Thus, $\sigma \dentailsn_{I\,}
\false$.  By robustness, it follows that $\KB_i \dentailsn_{I\,}
\false$.
But this can happen only if $\KB \sat \false$, which implies that $\KB
\sat \alpha \le 
\Pr(S) \le \beta$, contradicting our original assumption.
\eprf

\subsection{Proofs for Section~\protect{\ref{sec:rep-ind}}}\label{sec4prfs}

To prove Lemma~\ref{char.faithful}, it is useful to first prove
two additional results:

\lem\label{char.embed}
If $f$
is an $X$-$Y$ embedding, then $f(X) = Y$ and $f(\emptyset) = \emptyset$.
\elem

\prf Suppose that $f$ is an $X$-$Y$ embedding.  We first show that
$f(\emptyset) = \emptyset$.  From the definition of embedding, it follows
that $f(\emptyset) = f(X \inter \emptyset) = f(X) \inter f(\emptyset)$.
Thus, $f(\emptyset) \subseteq f(X)$.  But the definition of embedding also
implies that $f(\emptyset) = f(\overline{X}) = \overline{f(X)}$.
Thus, we have $\overline{f(X)} \subseteq f(X)$.  This can happen only if
$f(X) = Y$ and $f(\emptyset) = \overline{f(X)} = \emptyset$. \eprf

\lem\label{lem:correspondence}
If $f$ is a faithful $X$-$Y$ embedding, then
\begin{enumerate}\denselist
\item[(a)]
for any $\mu \in \Delta_X$, there is a measure $\nu \in \Delta_Y$ such
that $\nu$ corresponds to $\mu$;
\item[(b)]
for any $\nu \in \Delta_Y$, there is a measure $\mu \in \Delta_X$ such
that $\mu$ corresponds to $\nu$.
\end{enumerate}
\elem

\prf
To prove (a), consider the algebra of subsets of $Y$ the form $f(S)$,
for $S \in \F_X$.  Define a function $\nu'$ on the algebra via $\nu'(f(S)) =
\mu(S)$.  This mapping is well defined, for if $f(S) = f(T)$, then
faithfulness guarantees that $S = T$.  Moreover, $\nu'$ is a probability
measure on the algebra.  To see this, note by Lemma~\ref{char.embed}
that $\nu'(Y) = \nu'(f(X)) = \mu(X) = 1$.  Moreover, if $f(S) \inter f(T) =
\emptyset$, then (by definition of embedding) $f(S \inter T) = \emptyset$
and so, since $f$ is faithful, $S \inter T = \emptyset$ (for otherwise $f(S
\inter T) = f(\emptyset)$ by Lemma~\ref{char.embed}, but $S \inter T \ne
\emptyset$).  Thus, 
\[
\nu'(f(S) \union f(T)) = \nu'(f(S \union T)) = \mu(S \union T) = \mu(S) 
  + \mu(T) = \nu'(f(S)) + \nu'(f(T)).
\]
As shown by Horn and Tarski \citeyear{HT48}, it is possible to extend
$\nu'$ to a probability measure $\nu$ on $\F_Y$.%
\footnote{It is critical for this result that we are working with
finitely additive measures.  There  may not be a countably additive
measure $\nu$ extending $\nu'$, even if $\nu'$ is countably additive.
For example, take $\F_Y'$ to be 
the Borel sets on $[0,1]$ and take $\F_Y$ to be all subsets of $[0,1]$.
Let $\nu'$ be Lebesgue measure.  It is known that, under the continuum
hypothesis, there is no countably additive measure extending $\nu'$ defined on all subsets
of $[0,1]$ \cite{Ulam30} (see \cite{KT64} for further discussion).}
By construction, we have that $\nu$ corresponds to $\mu$. 

To prove (b), we use a very similar process.  Define a function $\mu$ on the
algebra of sets $S \subseteq X$ via $\mu(S) = \nu(f(S))$.  
It is easy to see that 
$\mu$ is already a probability measure in $\dists_X$,
which by construction corresponds to $\nu$.
\eprf

\commentout{
\medskip
We now must complete the proof of Lemma~\ref{char.embed} by showing that
that a probability measure $\nu'$ defined on a subalgebra $\F_Y'$ of
$\F_Y$ can be extended to a probability measure $\nu$ defined on all of
$\F_Y$.  In some cases, this is straightforward.  Define a
{\em basic\/} set in $\F_Y'$ to be a nonempty set none of whose nonempty
subsets are in $\F_Y'$.  (If all sets are measurable, then the basic sets
are just the singletons.)  Suppose that,  for every set $S$ in
$\F_Y'$, it is the case that $\nu'(S)$ is the sum of $\nu'(T)$ for
the basic sets $T$ contained in $S$.  (This will certainly be the case
if $S$ is finite, for example.)  For each basic set $S$, choose a
point $x_S \in S$.  For every set $T$ in $\F_Y$, define $\nu(T)$ as
$\sum_{x_S \in T} \nu'(S)$.  It is easy to check that $\nu$ is a measure
defined on $\F_Y$.  

However, this approach does not work if there is a set whose probability
is not the sum of the probability of the basic measurable sets it contains.  
Indeed, there  may not be a countably additive measure $\nu$ extending
$\nu'$, even if $\nu'$ is countably additive.  For example, take $\F_Y'$ to be
the Borel sets on $[0,1]$ and take $\F_Y$ to be all subsets of $[0,1]$.
Let $\nu'$ be Lebesgue measure.  It is known that, under the continuum
hypothesis, there is no countably additive measure extending $\nu'$ defined on all subsets
of $[0,1]$ \cite{Ulam30} (see \cite{KT64} for further discussion).  On
the other hand, if we require only finite 
additivity, it is always possible to extend, as the following result shows.

\lem\label{lem:extend}  If $\nu'$ is a (finitely additive) probability
measure defined on a subalgebra $\F_Y'$ of $\F_Y$, then there is a
probability measure $\nu$ defined on $\F_Y$ that extends $\nu'$.
\elem

\prf  Let $\A$ consist of all pairs $(\nu'',\F'')$
such that $\F''$ is an algebra, $\F_Y' \subseteq \F'' \subseteq \F_Y$,
and $\nu''$ is a measure on $\F''$ that extends $\nu'$. Define an order
on $\A$ by taking $(\nu_1, \F_1) \le (\nu_2, \F_2)$ if $\F_1
\subseteq \F_2$ and $\nu_2$ extends $\nu_1$.  We next show that every chain
(totally ordered sequence) $\{(\nu_\alpha,\F_\alpha): \alpha \in J\}$, of
such pairs (where $J$ is an arbitrary index set) 
has an upper bound.  Define $\F_\alpha^* = \union_{\alpha \in J} \F_\alpha$.
It is easy to check that $\F_\alpha^*$ is an algebra (however, note that
it is not in general a $\sigma$-algebra---it may not be closed under
countable unions).  If $U \in \F^*$, then $U \in \F_\beta$ for some
$\beta \in J$.  Define $\nu^*(U) = \nu_\beta(U)$.  Since
$\{(\nu_\alpha,\F_\alpha): \alpha \in J\}$ is a chain, the choice of
$\beta$ does not 
matter.  It is easy to check that $\nu^*$ is a measure that extends each
$\nu_\alpha$.  Thus, $(\nu_\alpha,\F_\alpha) \le (\nu^*,\F^*)$.  It
follows from Zorn's Lemma that there must be a pair $(\nu,\F^+) \in \A$ such
that $(\nu'',\F'') \le (\nu,\F^+)$ for all $(\nu'',\F'') \in \A$.  
We claim that $\F^+ = \F_Y$.

Suppose not.  
Choose $S \in \F_Y - \F^+$, and let $\F(S)$ be the smallest
subalgebra of $\F_Y$ containing $S$ and $\F^+$.  
Clearly $\F(S) \subseteq \F_Y$.  Let $\nu_S$ be a measure
on $\F(S)$ extending $\F_Y$.  It is well known that such a 
measure exists (see \cite{Rusp1} for a proof).  Then $(\nu_S,\F(S)) \in
\A$ and $(\nu,\F^+) < (\nu_S,\F(S))$.  This contradicts the choice of
$(\nu,\F^+)$.  Thus, $\F^+ = \F$, and we are done.  \eprf

\commentout{We first show how to extend $\nu'$ to $\F(S)$.  Define
$\nu^+(T \inter 
S) = \sup_{T' \subseteq T \inter S,\, T' \in \F_Y'} \nu'(T)$, and define
$\nu^+(T \inter \overline{S}) = \inf_{T' \supseteq T \inter
\overline{S},\, T' \in  F_Y'} \nu'(T)$.  Extend $\nu^+$ to all of
$\F(S)$ by additivity.  We need to show that $\nu^+$ is a measure and
that it extends $\nu$.  Note that for $T \in \F(S)$, we have $\nu^+(T) = 
\nu^+(T \inter S) + \nu^+(T \inter \overline{S})$.  Fix $\epsilon > 0$.
By definition, there must be some $U \in \F$ such that $U \subseteq T
\inter S$ and $\nu^+(T \inter S) \le \nu'U) + \epsilon$.  Moreover,
$\overline{U} \inter T \in \F$ and $\overline{U} \inter T \supseteq T
\inter \overline{S}$, so $\nu^+(T \inter \overline{S}) \le \nu'\overline{U}
\inter T)$.  Since $U \subseteq T, it follows that $U \union
(\overline{U} \inter T) = T$.
Thus, $$\nu^+(T \inter S) + \nu^+(T \inter \overline{S}) \le
\nu'U) + \nu'\overline{U} \inter T) + \epsilon = \nu'T) + \epsilon.$$
Since this is true for all $\epsilon > 0$, it follows that $\nu^+(T
\inter S) + \nu^+(T \inter \overline{S}) \le \nu'T)$.  
For the opposite
inequality, again fix $\epsilon$ and choose $U \in \F$ such that $U
\supseteq T \inter \overline{S}$ and $\nu^+(T \inter \overline{S}) \ge
\nu'U) - \epsilon$.  Thus, 
$$\nu^+(T \inter S) + \nu^+(T \inter \overline{S}) \ge
\nu'\overline{U} \inter T) + \nu'U) - \epsilon = \nu'T) - \epsilon.$$
Again, since this is true for all $\epsilon > 0$, it follows that 
$\nu^+(T \inter S) + \nu^+(T \inter \overline{S}) \ge \nu'T)$.  
Thus, $\nu^+(T \inter S) + \nu^+(T \inter \overline{S}) = \nu'T)$, as desired.

This shows that $\nu^+$ extends $\nu'$.  To show that $\nu^+$ is a
measure, it remains to show that if $U$ and $U'$ are disjoint subsets of 
$\F(S)$, then $\nu^+(U \union U') = \nu^+(U) + \nu^+(U')$.  We consider
here the case that $U = T \inter S$ and $U' = T' \inter S$.  The other
cases are similar (or simpler).  Fix $\epsilon > 0$.  By assumption,
there exist $V \in \F$ and $V' \in \F$ such that $V \subseteq T \inter
S$, $V' \subseteq V \inter S$, $\nu'(V) \ge \nu^+(T \inter S) -
\epsilon/2$, and $\nu'(V') \ge \nu^+(T' \inter S)$.  Then $V \union V'
\in \F'$, $V \union V' \subseteq (T \union T') \inter S$, and 
$\nu'(V \union V') \ge \nu^+(T \inter S) + \nu^+(T' \inter S) -
\epsilon$.  Thus, $\mu^+((T \union T') \inter S) \ge \nu^+(T \inter S) +
\nu^+(T' \inter S)$.  For the opposite inequality, ...}
}

We can now prove Lemma~\ref{char.faithful}.

\olem{char.faithful}
An $X$-$Y$ embedding $f$ is faithful if and only if for all
constraints $\KB$ and
$\theta$, we have $\KB \sat \theta$ iff $f^*(\KB) \sat f^*(\theta)$.
\eolem
\prf Suppose that $f$ is faithful.  To show that $\KB \sat \theta$ iff
$f^*(\KB) \sat f^*(\theta)$, we must show that
$\intension{\KB}{X} \subseteq \intension{\theta}{X}$ iff
$\intension{f^*(\KB)}{Y} \subseteq \intension{f^*(\theta}{Y}$.  The
``only if'' direction is immediate from the definition of $f^*$.  To
prove the ``if'' direction, suppose not.  Then there must exist some
$\mu \in \intension{\KB}{X} - \intension{\theta}{X}$ such that $f^*(\mu)
\subseteq \intension{f^*(\theta)}{Y}$.  Let $\nu$ be some probability
measure that corresponds to $\mu$.  Since $\nu \in f^*(\mu) \subseteq
f^*(\theta)$, there must be some $\mu' \in \intension{\theta}{X}$ such that
$\nu \in f^*(\mu')$.  Since $\mu' \ne \mu$, there must be some $S \in 
\F_X$ such that $\mu'(S) \ne \mu(S)$.  
Since $\nu \in f^*(\mu) \inter f^*(\mu')$, we must have both $\nu(f(S)) =
\mu(S)$ and $\nu(f(S)) = \mu'(S)$.  But this is a contradiction.
This completes the proof of the ``if'' direction.

For the converse, suppose we have $\KB \sat \theta$ iff $f^*(\KB) \sat
f^*(\theta)$ for all $\KB$ and $\theta$.  Given $S, T \in \F_X$, we
have the following chain of equivalences:

\begin{tabular}{ll}
&$S \subseteq T$\\
iff &$\Pr(S) = 1 \sat \Pr(T) = 1$\\
iff &$f^*(\Pr(S) = 1) \sat f^*(\Pr(T) = 1)$ (by assumption)\\
iff &$\Pr(f(S)) = 1 \sat \Pr(f(T)) = 1$ (by definition of $f^*$)\\
iff &$f(S) \subseteq f(T)$.
\end{tabular}

Thus, $f$ is faithful. \eprf

\medskip

\opro{pro:correspond1}
Let $f$ be a faithful $X$-$Y$ embedding.  Then
the following statements are equivalent:  
\begin{itemize}
\item[(a)] $\mu$ and $\nu$ correspond
under $f$; 
\item[(b)] for all formulas $\theta$, $\mu \sat\theta$ iff $\nu \sat
f^*(\theta)$.
\end{itemize}
\eopro
\prf
We first show that (a) implies (b).  
So suppose that $\mu$ and $\nu$ correspond under $f$.  
The only if direction 
of (b)
is trivial:
If $\mu \sat \theta$ then $\nu \in f^*(\mu) \subseteq f^*(\theta)$, 
since $f$ is faithful.
For the if direction, 
we proceed much as in the proof of Lemma~\ref{char.faithful}.
Assume that $\nu \sat f^*(\theta)$ but that $\mu
\not\sat \theta$.  Since $\nu \in f^*(\theta)$, by definition of $f^*$
there must be some $\mu' \in \intension{\theta}{X}$ such that $\nu \in
f^*(\mu')$.  Since 
$\mu' \sat \theta$ whereas $\mu \not\sat \theta$, we must have $\mu \neq
\mu'$.  Hence, there must be some $S \in \F_X$ such that $\mu(S) \neq
\mu'(S)$.   
Since $\nu \in f^*(\mu) \inter f^*(\mu')$, 
it follows that $\nu(f(S)) = \mu(S)$ and that $\nu(f(S)) =
\mu'(S)$, which gives the desired contradiction.  

We now show that (b) implies (a).  Assume by contradiction that $\mu$ and
$\nu$ do not correspond under $f$.  Then there must be some event $S
\in \F_X$ such that $\mu(S) \neq \nu(f(S))$.  Let $p = \mu(S)$ and let
$\theta$ be the constraint $\Pr(S) = p$.  Then $\mu \sat \theta$, whereas
$\nu \not\sat f^*(\theta)$, providing the desired contradiction.
\eprf

\medskip

\othm{robust} If an $\X$-inference procedure is robust
that satisfies DI2, DI4, and DI5,
then it is
representation independent. \eothm

\prf  Suppose that $\{I_X: X \in \X\}$ is a robust $\X$-inference procedure.
We want to show that it is representation independent.  So suppose $\KB,
\phi$ are constraints on $\dists_X$ and $f$ is an $X$-$Y$ embedding, for
some $X, Y \in \X$.  We want to 
show that $\KB \dentailsn_{I_X} \phi$ iff $f^*(\KB) \dentailsn_{I_Y}
f^*(\phi)$. 

\commentout{
We can assume without loss of generality that $X$ and $Y$ are disjoint.
(For, if not, find another space $Y'$ such that $|Y'| = |Y|$ and both
$X \inter Y' = \emptyset$ and $Y \inter Y' = \emptyset$.
Let $g$ be some isomorphism from $Y$ to $Y'$.  Note that $f\circ g$ is
an $X$-$Y'$ embedding, and $g^{-1}$ is a $Y'$-$Y$ embedding. If we can
prove the result for disjoint spaces then, using this result, we have
$\KB \dentailsn_I \phi$ iff $(f\circ g)^*(\KB) \dentailsn_I
(f \circ g)^*(\phi)$ iff $(f \circ g \circ g^{-1})^*(\KB) \dentailsn_I
(f \circ g \circ g^{-1})^*(\phi)$ iff $f^*(\KB) \dentails f^*(\phi)$.)

Let $\psi$ be a constraint that characterizes the embedding $f$.
That is, $\psi$ is the conjunction of the constraints
$\Pr(S \dimp f(S)) = 1$, for each $S \subseteq X$.
We claim that $\psi$ is $X$-conservative over $\KB$ and $Y$-conservative
over $\KB^*$.   To show that $\psi$ is $X$-conservative over $\KB$, we
must show that $\proj_X(\intension{\KB \land \psi}{X
\times Y}) = \intension{\KB}{X}$.  Clearly $\intension{\KB}{X} \subseteq
\proj_X(\intension{\KB \land \psi}{X \times Y})$, since
if $\mu \sat \KB \land \psi$, then $\mu_X \sat \KB$.
For the opposite inclusion, suppose $\nu \in \intension{\KB}{X}$.
There is a measure $\mu$
on $X \times Y$ such that $\mu(S \times f(S)) = \nu(S)$ for each $S
\in \F_X$.
To see this, consider the algebra consisting of unions of sets of
the form $S \times f(T)$, where either $T=S$ or $T$ is disjoint from
$S$.  It is easy to check that this is indeed an algebra: the
intersection of two sets of this form is again of this form.  It is
also easy to see that there is a unique
measure $\mu$ on this algebra that satisfies $\mu(S \times f(S)) =
\nu(S)$ and $\mu(S \times f(T)) = 0$ if $T \inter S = \emptyset$.
Extend this measure arbitrarily to $X \times Y$.
Since $\mu(S \times Y) = \mu(S \times f(S)) + \mu(S \times
\overline{f(S)}) = \nu(S)$, it follows that $\mu_X = \nu$.
Moreover, since $\mu(S \dimp f(S)) = \mu(S \times f(S) \union
\overline{S} \times \overline{f(S)}) = \nu(S) + \nu(\overline{S}) = 1$
(since $f(\overline{S}) = \overline{f(S)}$), it follows that
$\mu \sat \psi$.  Thus, $\psi$ is $X$-conservative over $\KB$.

Similar arguments show that $\psi$ is $Y$-conservative over $f^*(\KB)$.
The inclusion $\intension{f^*(\KB)}{Y} \subseteq
\proj_Y(\intension{\KB \land \psi}{X \times Y})$ is once again
straightforward.  
Given a measure $\nu \in \intension{f^*(\KB)}{Y}$, we want to construct
a measure $\mu$ on $X \times Y$ such that $\mu \sat \psi$ and $\mu_Y =
\nu$. Define $\mu(x,y) = \nu(y)$ if $y \in f(x)$ and $\mu(x,y) = 0$ if $y 
\notin f(x)$.  Using the definition of embedding and
Lemma~\ref{char.embed}, we can easily show that there is a unique
$x$ such that 
$f(x) \ni y$.  
Thus, $\mu(X \times \{y\}) = \nu(y)$,
so $\mu_Y = \nu$.  Moreover, it is easy to see that $\mu(S \times f(S))
= \nu(f(S))$, so similar arguments to those above show that $\mu(S \dimp
f(S)) = 1$.  Thus, $\mu \sat \psi$, 
showing the other side of the inclusion, and thereby that
$\psi$ is $Y$-conservative over $f^*(\KB)$.

It is easy to see that $\sat \psi \rimp (\sigma \dimp f^*(\sigma))$ for
all constraints $\sigma$ on $\dists_X$,
since $\psi$ captures the embedding.  This means (among other things)
that $\sat \KB \land \psi \dimp f^*(\KB) \land \psi$.
Using robustness, we get that $\KB \dentailsn_I \phi$ iff
$\KB \land \psi \dentailsn_I \phi$ iff $f^*(\KB) \land\psi \dentailsn_I
f^*(\phi)$ iff $f^*(\KB) \dentailsn_I f^*(\phi)$.
}%
Let $\psi$ be the following constraint on $\dists_{X \times Y}$:
$$(\phi \dimp f^*(\phi)) \land  (\KB \dimp f^*(\KB)).$$
We claim that $\psi$ is $X$-conservative over $\KB$ and
$Y$-conservative over $f^*(\KB)$.  Thus, we must show that 
$\proj_X(\intension{\KB \land \psi}{X \times Y}) =
\intension{\KB}{X}$ and 
$\proj_Y(\intension{f^*(\KB) \land \psi}{X \times Y}) = 
\intension{f^*(\KB)}{Y}$.  
We show that $\proj_X(\intension{\KB \land \psi}{X \times Y}) =
\intension{\KB}{X}$ here; the argument that
$\proj_Y(\intension{f^*(\KB) \land \psi}{X \times Y}) = 
\intension{f^*(\KB)}{Y}$ is almost identical.

Clearly if $\mu \in
\intension{\KB \land \psi}{X \times Y}$ then $\mu_X \in
\intension{\KB}{X}$, so $\proj_X(\intension{\KB \land \psi}{X \times Y})
\subseteq \intension{\KB}{X}$.
  For the opposite inclusion, suppose that $\nu \in
\intension{\KB}{X}$.  We want to find a measure $\nu' \in 
\intension{\KB \land \psi}{X \times Y})$ such that $\nu'_X = \nu$.
Let $\nu''$ be any measure in $f^*(\mu)$ and let
$\nu' \in \dists_{X \times Y}$ be the ``crossproduct'' of $\nu$ and
$\nu''$; that is, $\nu'(A \times B) = \nu(A) \nu''(B)$.  Clearly
$\nu'_X = \nu$.  To see that $\nu' \in \intension{\KB \land \psi}{X
\times Y})$, it clearly suffices to show that $\nu' \sat \psi$.  But
since $\nu$ and $\nu''$ correspond under $f$, it is immediate from
Proposition~\ref{pro:correspond1} that $\nu \sat \KB$ iff $\nu'' \sat
f^*(\KB)$ and $\nu \sat \phi$ iff $\nu'' \sat f^*(\phi)$.  Thus, $\nu
\sat \psi$, as desired.  

Now suppose that $\KB \dentailsn_{I_X\,} \phi$.
By DI2 and DI5, $\KB \land \psi$ is in the domain of $I_{X \times Y}$.
By robustness, $\KB \land \psi \dentailsn_{I_{X \times Y}\,} \phi$.
Thus, $I(\intension{\KB \land \psi}{X \times Y}) \subseteq \intension{\phi}{X
\times Y}$.  Since $I(\intension{\KB \land \psi}{X \times Y}) \subseteq
\intension{\KB \land \psi}{X \times Y} \subseteq \intension{\phi \dimp
f^*(\phi)}{X \times Y}$, it follows that 
$I(\intension{\KB \land \psi}{X \times Y}) \subseteq \intension{f^*(\phi)}{X
\times Y}$.  Moreover, $\KB \land \psi$ is equivalent to
$f^*(\KB) \land \psi$, so $I(\intension{f^*(\KB) \land \psi}{X \times Y})
\subseteq \intension{f^*(\phi)}{X \times Y}$, \ie $f^*(\KB) \land \psi
\dentailsn_{I_{X \times Y}\,} f^*(\phi)$.  
By DI4, $f^*(\KB)$ is in the domain of $I_Y$.
Since $\psi$ is
$Y$-conservative over $f^*(\KB)$, the robustness of $\{I_X: X \in \X\}$ 
implies that $f^*(\KB)
\dentailsn_{I_Y\,} f^*(\phi)$.  The opposite implication (if $f^*(\KB)
\dentailsn_{I_Y\,} f^*(\phi)$ then $\KB \dentailsn_{I_X\,} \phi$) goes the
same way.  Thus, $\{I_X: X \in \X\}$ is representation independent.
\eprf

Next, we turn our attention to Theorems~\ref{almosttrivial1}
and~\ref{noindep}.  Both of these results follow in a relatively
straightforward way from one key proposition.  Before we state it, we need
some definitions. 

\dfn We say that a constraint $\KB$ on $\dists_X$ {\em depends only
on $S_1, \ldots, S_k \in \F_X$\/} (the sets $S_1, \ldots, S_k$ are
not necessarily disjoint) if, whenever $\mu, \mu' \in \dists_X$ agree on
$S_1, \ldots, S_k$, then $\mu\sat \KB$ iff $\mu' \sat \KB$. \edfn

For example, if $\KB$ has the form $\Pr(S_1) > 1/3 \land \Pr(S_2) \le
3/4$, then $\KB$ depends only on $S_1$ and $S_2$.  Similarly,
if $\KB$ has the form $\Pr(S_1 \mid S_2) > 3/4$, then $\KB$ depends only on
$S_1$ and $S_2$.

\dfn Given $S_1, \ldots, S_k \in \F_X$, an {\em atom\/} over
$S_1, \ldots, S_k$ is a set of the form $T_1 \inter \ldots \inter T_k$,
where $T_i$ is either $S_i$ or $\overline{S_i}$. \edfn

\pro\label{notrepind} Suppose that $\{I_X: X \in \X\}$ is an $\X$-inference
procedure 
and, for some $X \in \X$, there exist 
$S, S_1, \ldots, S_K \in \F_X$
and a consistent constraint $\KB$ on $\dists_X$ that depends only on $S_1,
\ldots, S_k$, such that the following two conditions
are satisfied:
\begin{itemize}
\item
both $T \inter S$ and $T \inter \overline{S}$ are nonempty
for every nonempty atom $T$ over $S_1, \ldots, S_k$,
\item $\KB \dentailsn_{I_X\,} \alpha
< \Pr(S) < \beta$, where either $\alpha > 0$ or $\beta < 1$.
\end{itemize}
Then $\{I_X: X \in \X\}$ is not representation independent. \epro

\prf Suppose, by way of contradiction, that $\{I_X: X \in \X\}$ is a
representation-independent inference procedure but nevertheless, for
some $X \in \X$, there exists sets $S, S_1, \ldots, S_k \in \F_X$ and
a knowledge base $\KB$ that satisfies the conditions above, for some
$\alpha, \beta$.
Assume that $\alpha > 0$
(a similar argument can be used to deal
with the case that $\beta < 1$).

Let $T_1, \ldots, T_M$ be the nonempty atoms over $S_1, \ldots, S_k$.
Choose $N$ such that $1/N < \alpha$.  Our goal is to find a collection
$f_1, \ldots, f_N$ of embeddings of $X$ into some $Y \in \X$ such that
each of these embeddings has the same effect on $\KB$, but such that
the sets $f_j(S)$ are disjoint.  
Since $\KB \dentailsn_{\infp_X\,} \Pr(f_j(S)) > \alpha$ for $j = 1, \ldots,
N$, and $f_j^*(\KB) = f^*(\KB)$ for $j = 1, \ldots, N$, 
it will follow that
$f^*(\KB) \dentailsn_{\infp_Y\,} \Pr(f_j(S)) > \alpha$ for $j = 1, \ldots, N$,
a contradiction.  We proceed as follows.

By assumption, there exists a set $Z$ in $\X$ such that $|Z| = MN$.
Let $Y = X \times Z$.  Since $\X$ is closed under crossproducts, $Y \in \X$.
Suppose that $Z = \{z_1, \ldots, z_{MN}\}$, and let 
$Z_i = \{z_{N(i-1) + 1}, \ldots, z_{Ni}\}$, for $i = 1, \ldots, M$.
Thus, the $Z_i$s partition $Z$ into $M$ disjoint sets, each of
cardinality $N$. Let $B_i = X \times Z_i$, and let $B_{ij} = X \times
\{z_{N(i-1) + j}\}$, for $j = 1, \ldots, N$.
It is easy to see that we can find faithful
$X$-$Y$ embeddings $f_1, \ldots, f_N$ such that
\begin{enumerate}
\item $f_j(T_i) = B_i$, for $i = 1, \ldots, M$, $j = 1, \ldots, N$,
\item $f_j(T_i \inter S) = B_{ij}$, for $i = 1, \ldots, M$, $j = 1,
\ldots, N$.
\end{enumerate}
Notice that we need the assumption that both $T_i \inter S$ and
$T_i \inter \overline{S}$ are nonempty for $T_1, \ldots, T_M$
(that is, for each nonempty atom over $S_1, \ldots, S_k$) to
guarantee that we can find such faithful embeddings.  
For if $T_i \inter S = \emptyset$, then since $f_j$ is an embedding, $f(T_i
\inter S) = \emptyset \ne B_i$; and if $T_i \inter \overline{S} =
\emptyset$, then $f_j(T_i \inter \overline{S}) = f_j(T_i) - f(T_i \inter S))
= \emptyset$, which means that $B_i = B_{ij}$, again inconsistent with the
construction.  

It is easy to check that, since $\KB$ depends only on $S_1, \ldots, S_k$,
$f_j^*(\KB)$ depends only on $f_j(S_1), \ldots, f_j(S_k)$, for $j = 1,
\ldots, N$.  We next show that $f_j(S_i)$ is independent of $j$; that
is, $f_j(S_i) = f_{j'}(S_i)$ for $1 \le j, j' \le N$.
Notice that for $h = 1, \ldots, k$, we have that $f_j(S_h) = \union_{T_i
\subseteq S_h} f_j(T_i) = \union_{\{i: T_i \subseteq S_h\}} B_i$.  Thus,
$f_j(S_h)$ is independent of $j$, as desired.  Since 
$f_j^*(\KB)$ depends
only on $f_j(S_1), \ldots, f_j(S_k)$, it too must be independent of $j$.
Let $\KB^*$ be
$f_1^*(\KB)$ (which, as we have just observed, is identical to $f_2^*(\KB),
\ldots, f_k^*(\KB)$).

Since, by assumption, $\{I_X: X \in \X\}$ is representation independent, and
$\KB \dentailsn_{\infp_X\,} Pr(S) > \alpha$, we have that $\KB^*
\dentailsn_{\infp_Y\,} \Pr(f_j(S)) > \alpha$, for $j = 1, \ldots, N$.  Thus,
$\KB^* \dentailsn_{\infp_Y\,} \Pr(f_1(S)) > \alpha \land \ldots \land
\Pr(f_N(S)) > \alpha$.  But
note that, by construction, $f_j(S) = \union_{\{i: T_i \inter S \ne
\emptyset\}} B_{ij}$.  Thus, the sets $f_j(S)$ are pairwise disjoint.
Since $\alpha > 1/N$, we cannot have $N$ disjoint sets each with
probability greater than $\alpha$.  Thus, $\KB^* \dentailsn_{\infp_Y\,}
\false$.  But $\KB$ is consistent, so $\KB^* = f_j(\KB)$ must be as
well. Thus, $I_Y(\KB^*) \ne \emptyset$, by assumption.  But this
contradicts our conclusion that $\KB^* \dentailsn_{I_Y\,} \false$.
Thus, $\{I_X: X \in \X\}$ cannot be representation independent. \eprf

We can use Proposition~\ref{notrepind} to 
help prove Theorem~\ref{almosttrivial1}.

\othm{almosttrivial1} 
If $\{I_X: X \in \X\}$ is a representation-independent $\X$-inference
procedure 
then, for all $X \in \X$,  $I_X$ 
is essentially entailment for 
all objective knowledge bases in its domain.
\eothm

\prf  Suppose, by way of contradiction, that $\{I_X: X \in \X\}$ is
representation independent but $I_X$ is not essentially entailment for
some $X \in \X$ and objective knowledge base $\KB$.  Then there must be
some set $S \in \F_X$ such that
$\KB \dentailsn_{I_X\,} \alpha < \Pr(S) < \beta$ and $\KB \not\sat \alpha \le
\Pr(S) \le \beta$.  Without loss of generality, we can assume that
$\KB$ has the form $\Pr(T) = 1$ for some $T \in \F_X$.  Moreover, we
can assume 
that if $\overline{T} \ne \emptyset$, then $\overline{T}$ has a
nonempty, measurable strict subset.  
(For otherwise, choose $Y = \{y,y'\} \in \X$ and
consider the space $X' = X \times Y$.  By assumption, $X' \in \X$.
Let $f$ be the $X$-$Y$ embedding
that maps $U \in \F_X$ to $U \times Y$.  Since $I$ is representation
independent, we have that $\Pr(T \times Y) = 1 \dentailsn_{I\,} \alpha < 
\Pr(S \times Y) < \beta$, and 
$\overline{T} \times \{y\} \subset \overline{T} \times Y$.)

If $\overline{T}$ is nonempty, let $Z$ be any nonempty, 
measurable strict subset of $\overline{T}$ (which exists by assumption);
otherwise let $Z$ be the empty set.  
Let $U$ be the set $(T
\inter S) \union (\overline{T} \inter Z)$.
Notice that $S \inter T = U \inter T$.  Moreover, since, for any set
$V$,
$\Pr(T) = 1 \rimp \Pr(V) = \Pr(V \inter T)$ is valid,
it follows from
Reflexivity and Right Weakening that
$\KB \dentailsn_{I_X\,} \Pr(V) = \Pr(V \inter T)$.
Thus, $\KB \dentailsn_{I_X\,} \Pr(S) = \Pr(S \inter T) =
\Pr(U \inter T) = \Pr(U)$.  It follows that $\KB \dentailsn_{I_X\,}
\alpha < \Pr(U) < \beta$.

We now want to apply Proposition~\ref{notrepind}.  Note that $\KB$
depends only on $T$.   Thus, we must show that
$T \inter U$ and $T \inter \overline{U}$ are nonempty, and
if $\overline{T}$ is nonempty, then $\overline{T} \inter
U$ and $\overline{T} \inter \overline{U}$ are as well. As we observed
above, $T \inter U = T \inter S$.  Thus, if $T \inter U = \emptyset$,
then $T \subseteq \overline{S}$, contradicting our assumption that $\KB
\dentailsn_I \Pr(S) > 0$.  It is easy to see that $T \inter \overline{U}
= T \inter \overline{S}$.  Again, we cannot have $T \inter \overline{U}
= \emptyset$, for then
$T \subseteq S$, contradicting our assumption that $\KB
\dentailsn_I \Pr(S) < 1$.  By construction, $\overline{T} \inter U  =
\overline{T}
\inter Z = Z$.  By assumption, if $\overline{T} \ne \emptyset$, then $Z
\ne \emptyset$.   Finally, $\overline{T} \inter \overline{U} =
\overline{T} \inter \overline{Z}$; again, by construction, this is a
nonempty set if $\overline{T} \ne \emptyset$.
It now follows from Proposition~\ref{notrepind}
that $\{I_X: X \in \X\}$ is not representation independent.  \eprf

\ocor{almosttrivial3}
If $\{I_X: X \in \X\}$ is a representation-independent $\X$-inference
procedure, then 
for all $X \in \X$, 
if $\KB$ is an objective knowledge base putting constraints on
$\dists_X$, and $\KB \dentailsn_{I_X\,} \alpha < \Pr(S) < 
\beta$ for some $\alpha \ge 0$ and $\beta \le 1$, then $\alpha = 0$ and
$\beta = 1$. 
\eocor
\prf Assume the hypotheses of the corollary hold.  Since $\KB$ is
objective, it is of the form $\Pr(T) = 1$ for some $T \in \F_X$.  Then there
are three possibilities.  Either (1) $T \subseteq S$, (2) $T \subseteq
\overline{S}$, or (3) both $T \inter S$ and $T \inter \overline{S}$ are
nonempty.  If (1) holds, we have $\KB \sat \Pr(S) =1$, while if (2)
holds, we have $\KB \sat \Pr(S) = 0$.  Thus, both (1) and (2) are
incompatible with $\KB \dentailsn_{I_X\,} \alpha < \Pr(S) < \beta$.  On the
other hand, if (3) holds, it is easy to see that for all $\gamma$,
$\Pr(S) = \gamma$ is consistent with $\KB$ (since there is a probability
measure that assigns probability $\gamma$ to $T \inter S$ and
probability $1- \gamma$ to $T \inter \overline{S}$).  
Since $\KB \dentailsn_{I_X\,} \alpha < \Pr(S) < \beta$, by
Theorem~\ref{almosttrivial1}, we must have $\KB \sat \alpha \le \Pr(S)
\le \beta$.  It follows that the only choices of $\alpha$ and $\beta$
for which this can be true are $\alpha = 0$ and $\beta = 1$. \eprf

\othm{noindep} 
If $\{I_X: X \in \X\}$ is an $\X$-inference procedure that 
enforces minimal default independence
and satisfies DI1,
then $I_X$ is not representation independent.
\eothm

\prf Suppose that $\{I_X: X \in \X\}$ is an $\X$-inference procedure that
enforces minimal default independence
and satisfies DI1.
Choose $X = \{x,x'\} \in \X$
and let $\KB$ be $1/3 \le \Pr(x) \le 2/3$.
By assumption, $X \times X \in \X$.
We can view $\KB$ as
a constraint on $\dists_{X \times X}$; in this case,
it should be interpreted
as $1/3 \le \Pr(\{x\} \times X) \le 2/3$.
by DI1, $\KB$ is is the domain of $I_{X \times X}$.
Note that $\KB$ is equivalent to the constraint $1/3 \le \Pr(x') \le 2/3$.
By minimal default independence, we have
that $\KB \dentailsn_{I_{X \times X\,}} \Pr((x,x)) >  \Pr(x \times X)/3$
and that $\KB 
\dentailsn_{I_{X \times X\,}} \Pr((x',x')) > \Pr(x' \times X)/3$.  
Applying straightforward probabilistic reasoning, we get that
$\KB \dentailsn_{I_{X \times X}\,} \Pr(\{(x,x),(x',x')\}) > 1/3$.
We now apply Proposition~\ref{notrepind}, taking $S$ to be
$\{(x,x),(x',x')\}$ and $S'$ to be $\{(x,x),(x,x')\}$.  Note that $\KB$
depends only on $S'$.  It is almost immediate from the definition of $S$
and $S'$ that
all of $S \inter S'$, $\overline{S} \inter S'$, $S \inter
\overline{S'}$, and $\overline{S} \inter \overline{S'}$ are nonempty.
Thus, by Proposition~\ref{notrepind}, $\{I_X: X \in \X\}$ is not
representation independent.  
\eprf

\medskip

\olem{I*} 
Let $\X$ consist of only countable sets.
Then $\{\infp^0_X: X \in \X\}$ is a
representation-independent $\X$-inference procedure. 
\eolem

\prf
As we said in the main part of the text,
it easily follows from Proposition~\ref{choosedists.side2} that
$\infp^0_X$ is an 
inference procedure for all $X \in \X$, since it is easily seen to have
the five properties 
described in the proposition.  To see
that $\infp^0$ is representation independent, suppose that $f$ is
a faithful $X$-$Y$ embedding, for $X, Y \in \X$.  Clearly $\KB$ is
objective if and only if 
$f^*(\KB)$ is objective.  If $\KB$ is not objective, then it is easy to
see that 
$\KB \dentailsn_{\infp^0\,} \theta$ iff $f^*(\KB) \dentailsn_{\infp^0\,}
f^*(\theta)$, since $\dentailsn_{\infp^0}$ reduces to entailment in this
case.  So suppose that $\KB$ is objective
and has the form $\Pr(T) = 1$, for some $T \in \F_X$.
Then $\KB
\dentailsn_{\infp^0\,} \theta$ iff $\KB \land \KB^+ \sat \theta$.
By Lemma~\ref{char.faithful}, this holds iff $f^*(\KB) \land f^*(\KB^+)
\sat f^*(\theta)$.  On the other hand,
$f^*(\KB) \dentailsn_{\infp^0\,} f^*(\theta)$ iff $f^*(\KB) \land
(f^*(\KB))^+ \sat f^*(\theta)$
Thus, it suffices to show that 
$f^*(\KB) \land f^*(\KB^+) \sat f^*(\theta)$ iff $f^*(\KB) \land
(f^*(\KB))^+ \sat f^*(\theta)$.  
It is easy to show that $(f^*(\KB))^+$ implies $f^*(\KB^+)$, so that if
$f^*(\KB) \land f^*(\KB^+) \sat f^*(\theta)$ then $f^*(\KB) \land
(f^*(\KB))^+ \sat f^*(\theta)$.  It is not necessarily the case that
$f^*(\KB^+)$ implies $(f^*(\KB))^+$.  For example, consider the
embedding described in Example~\ref{embed.color}.  In that case, 
if $\KB$ is the objective knowledge base $\Pr(\colorful) = 1$, $\KB^+$
is empty, and hence so is $f^*(\KB^+)$, while $(f^*(\KB))^+$ includes
constraints such as $0 < \Pr(\green) < 1$.  Nevertheless, suppose that
$f^*(\KB) \land (f^*(\KB))^+ \sat f^*(\theta)$ and, by way of
contradiction, there is some $\nu$ such that $\nu \sat f^*(\KB) \land
f^*(\KB^+) \land \neg f^*(\theta)$.  Choose $\mu$ such that $\nu \in
f^*(\mu)$.  Then $\mu$ and $\nu$ correspond, so $\mu \sat \KB \land
\KB^+ \land \neg \theta$.  It is easy to show that
there exists $\nu' \in
f^*(\mu)$  such that $0 < \nu'(S) < 1$ for all nonempty subsets
of $S$ of $f(T)$.  To see this, note that if $\mu(x) \ne 0$, then it
suffices to ensure that $\nu'(f(x)) = \mu(x)$ and $\nu'(y) \ne 0$ for all
$y$ in $f(x)$.  Since $Y$ is countable, this is straightforward.  
Since $\mu$ and $\nu'$ correspond, we must have that $\nu' \sat
\neg f^*(\theta) \land f^*(\KB)$.  By construction, $\nu' \sat
(f^*(\KB))^+$.  This contradicts the assumption that $f^*(\KB) \land
(f^*(\KB))^+ \sat f^*(\theta)$. \eprf

\medskip

\olem{I1} 
Suppose that $\X$ consists only of measure spaces of the form $(X,2^X)$,
where $X$ is finite.  Then  $\{I^1_X: X \in \X\}$ 
is a representation-independent $\X$-inference
procedure.
\eolem

\prf Suppose that $X, Y \in \X$, $\KB$ and $\phi$ are constraints on
$\dists_X$, and $f$ is an $X$-$Y$ embedding.  We must show that
$\KB \dentailsn_{I^1_X\,} \phi$ iff $f^*(\KB) \dentailsn_{I^1_Y\,} f^*(\phi)$.
For the purposes of this proof, we say that a subset $A$ of $\dists_X$
is {\em interesting\/} if
there exists some $S \in \F_X$ such that
$A = \{\mu \in \dists_X : \mu(S) \ge 1/4\}$.
It is easy to see that if $\KB$ is interesting then $f^*(\KB)$ is
interesting.   The converse is also true, given our assumption that $\X$
consists of only finite spaces where all sets are measurable.  
For suppose that $f^*(\KB)$ is interesting.  Then there is a
set $T \subseteq Y$ such that $f^*(\KB) = \{\nu \in \dists_Y: \nu(T) \ge
1/4\}$. Let $\A = \{S' \subseteq X: f(S') \supseteq T\}$.  Since $X$ is
finite, so is $\A$; it easily follows that $S = \inter \A \in \A$.%
\footnote{This is not in general true if $X$ is infinite without 
the additional requirement that $f(\union_i\, A_i) = \union_i f(A_i)$
for arbitrary unions.}  
Clearly if $\mu(S) \ge 1/4$, then $f^*(\mu) \subseteq f^*(\KB)$, so $\mu \in
\intension{\KB}{X}$.  Thus, $\intension{\KB}{X} \supseteq \{\mu \in
\dists_X: \mu(S) 
\ge 1/4\}$.  On the other hand, if $\mu \in \KB$, then $f^*(\mu) \subseteq
f^*(\KB)$.  Thus, if $\nu \in f^*(\mu)$, since $S \in \A$, it must be
the case that $\mu(S) = \nu(f(S)) \ge \nu(T) \ge 1/4$.  Thus, 
$\intension{\KB}{X} \subseteq \{\mu \in
\dists_X: \mu(S) \ge 1/4\}$.  It follows that $\KB$ is equivalent to
$\Pr(S) \ge 
1/4$, and so must be interesting.  (We must also have $T = f(S)$,
although this is not needed for the result.)

If $\KB$ is not interesting, then $\KB \dentailsn_{I^1_X}
\phi$ iff $\KB \sat \phi$ iff $f^*(\KB) \sat f^*(\phi)$ (since
entailment is representation independent) iff $f^*(\KB) \dentailsn_{I^1_Y}
\phi$.  On the other hand, if $\KB$ is interesting, then $\KB$ is
equivalent to $\Pr(S) \ge 1/4$ for some $S \subseteq X$, and $f^*(\KB)$ is
equivalent to $\Pr(f(S)) \ge 1/4$.  Moreover, $\KB \dentailsn_{I^1_X \,} \phi$
iff $\Pr(S) \ge 1/3 \sat \phi$ iff $\Pr(f(S)) \ge 1/3 \sat f^*(\phi)$
iff $f^*(\KB) \dentailsn_{I^1_Y\,} \phi$.  Thus, we get representation
independence, as desired. \eprf

\subsection{Proofs for Section~\protect{\ref{limited}}}\label{sec6prfs}

\opro{correspond2}
Suppose that  $f$ is a faithful $X$-$Y$ embedding, $\D_X
\subseteq \dists_X$, and $\D_Y \subseteq \dists_Y$.  The following two
conditions are equivalent: 
\begin{itemize}
\item[(a)] $\D_X$ and $\D_Y$ correspond under $f$;
\item[(b)] for all $\theta$, $\D_X \sat \theta$ iff $\D_Y \sat 
f^*(\theta)$.
\end{itemize}
\eopro

\prf
To prove that (a) implies (b), assume by way of contradiction
that, for some $\theta$, $\D_X \sat \theta$ but $\D_Y \not\sat f^*(\theta)$.
Then there is some $\nu \in \D_Y$ such that $\nu \not\sat f^*(\theta)$.  Let
$\mu \in \D_X$ be a measure corresponding to $\mu$.   Then, by 
Proposition~\ref{pro:correspond1}, we have that $\mu \not\sat \theta$, the
desired contradiction.  The proof for the other direction of (a) is identical.

To prove that (b) implies (a), first consider a measure $\mu \in \D_X$.
We must find a $\nu \in \D_Y$ such that $\nu$ corresponds to $\mu$.
Suppose that $X = \{x_1, \ldots, x_n\}$ (recall that we are restricting to
finite spaces in Section~\ref{limited}) and that $\mu(x_i) = a_i$, $i =
1, \ldots, n$.  Let $\theta$ be the constraint $\land_{i=1}^n
\Pr(\{x_i\}) = a_i$.  By our assumptions about the language, this
constraint is in the language. 
Clearly $\intension{\theta}{X} = \{\mu\}$.  
Since $\mu
\in \D_X$, we know that $\D_X \not\sat \neg \theta$.  Hence, $\D_Y \not\sat
f^*(\neg \theta)$, so that there exists $\nu \in \D_Y$ such that $\nu \not\in
f^*(\neg \theta)$.  Hence $\nu \in f^*(\theta) = f^*(\{\mu\})$. 
By definition of $f^*$, $\nu$ corresponds to $\mu$.  

Now consider a measure $\nu \in \D_Y$, and let $\mu$
be the measure in $\Delta_X$ that corresponds to $\nu$.  
Assume by way of contradiction that $\mu \not\in \D_X$.  
Taking $\theta$ as above, it follows that 
$\D_X \sat \neg \theta$ and,
therefore, by assumption, $\D_Y \sat f^*(\neg \theta)$.
Thus, $\nu \sat f^*(\neg \theta)$.  But $\mu \sat \theta$ and, by
assumption, $\mu$ and $\nu$
correspond.  This contradicts Proposition~\ref{pro:correspond1}.
\eprf

\medskip

\othm{relative.entropy}
Let $\theta$ be an arbitrary constraint on $\dists_X$.  If 
$f$ is a faithful $X$-$Y$ embedding and
$\mu$ and
$\nu$ correspond under $f$, then $\mu|\theta$ and $\nu|f^*(\theta)$ also
correspond under $f$.
\eothm

\prf
Assume that $\mu$ and $\nu$ correspond under $f$.  Recall that we are
assuming in this section that $X$ is a finite space; let $X =
\{x_1,\ldots,x_n\}$.  Let $Y_i = f(x_i)$.    
Given any distribution $\nu'' \in \dists_Y$, define
$\nu''_i = \nu''|Y_i$ and let $(f^*)^{-1}(\nu'')$ denote the unique $\mu'' \in
\dists_X$ such that $\nu'' \in f^*(\mu'')$.  

Now suppose that  $\mu' \in \mu|\theta$.  Define $\nu' \in \dists_Y$ to be the
measure such that 
\[
  \nu'(y) = \mu'(x_i) \cdot \nu_i(y),
\]
where $i$ is the index such that $y \in Y_i$. 
Since $\nu_i = \nu|Y_i$, it follows that $\nu_i(Y_i) = 1$.  Thus, 
$\nu'(Y_i) = \mu(x_i)$, and $\nu'$ leaves the
relative probabilities of elements within each $Y_i$ the same as in $\nu$. 
It is easy to verify that $\nu'$ and $\mu'$ correspond.  Hence, by
Proposition~\ref{pro:correspond1}, $\nu'
\sat f^*(\theta)$.  We claim that $\nu' \in \nu|f^*(\theta)$.  To show that,
we need show only that $\KLD{Y}{\nu'}{\nu}$ is minimal among all
$\KLD{Y}{\nu''}{\nu}$ 
such that $\nu'' \sat f^*(\theta)$.  
It follows from standard properties of relative entropy \cite[Theorem
2.5.3]{CoverThomas} 
that for all $\nu'' \in \dists_Y$, we have
\begin{equation}\label{eq:decomp}
\KLD{Y}{\nu''}{\nu} = \KLD{X}{(f^*)^{-1}(\nu'')}{(f^*)^{-1}(\nu)} + 
   \sum_{i=1}^n \KLD{Y}{\nu''_i}{\nu_i}.
\end{equation}
Note that $\nu_i = \nu'_i$, so $\KLD{Y}{\nu_i'}{\nu_i} = 0$, for $i = 1,
\ldots, n$.  Thus, it follows from (\ref{eq:decomp}) that
$\KLD{Y}{\nu'}{\nu} = \KLD{X}{\mu'}{\mu}$.

Now, let $\nu''\in \dists_Y$ be such that
$\nu'' \sat f^*(\theta)$  and let $\mu'' = (f^*)^{-1}(\mu'')$.  
Since $\nu''$ and $\mu''$ correspond under $f$, it follows from
Proposition~\ref{pro:correspond1} that $\mu'' \sat \theta$.
Using~(\ref{eq:decomp}) once
again, we have that 
\begin{eqnarray*}
\KLD{Y}{\nu''}{\nu} &= & \KLD{X}{\mu''}{\mu} + 
   \sum_{i=1}^n \KLD{Y}{\nu''_i}{\nu_i} \\
&\ge & \KLD{X}{\mu''}{\mu}.
\end{eqnarray*}
But since $\mu' \in \mu|\theta$, we know that $\KLD{X}{\mu'}{\mu} \leq
\KLD{X}{\mu''}{\mu}$.  Hence we conclude that
\[
 \KLD{Y}{\nu''}{\nu} \geq  \KLD{Y}{\nu'}{\nu},
\]
so that $\nu' \in \nu|f^*(\theta)$.
\eprf

\bigskip

\othm{bootstrap}
If $f$ is a faithful $X$-$Y$ embedding, then 
$\infp^\P$ is invariant under $f$
iff $\P(X)$ and $\P(Y)$ correspond under $f$.
\eothm

\medskip
\prf Suppose that $f$ is a faithful $X$-$Y$ embedding.  By definition,
$\infp^\P$ is invariant under $f$ iff, for all $\KB$, $\theta$, we have
\begin{equation}\label{bootstrap1}
\mbox{$\KB \dentails_{\infp^\P\,} \theta$ iff $f^*(\KB) \dentails_{\infp^\P\,}
f^*(\theta)$.}
\end{equation}  
By definition of $\infp^\P$, (\ref{bootstrap1}) holds
iff
\begin{equation}\label{bootstrap2}
\mbox{$\P(X)|\KB \subseteq \intension{\theta}{X}$ iff $\P(Y)|f^*(\KB)
\subseteq \intension{f^*(\theta)}{Y}$ for all $\KB$, $\theta$.}
\end{equation}
By Proposition~\ref{correspond2}, (\ref{bootstrap2}) holds iff
$\P(X)|\KB$ and $\P(Y)|f^*(\KB)$ correspond for all $\KB$.  
By Corollary~\ref{cor:cross.entropy}, if $\P(X)$ and $\P(Y)$ correspond,
then $\P(X)|\KB$ and $\P(Y)|f^*(\KB)$ correspond for all $\KB$.
On the other hand, if $\P(X)|\KB$ and $\P(Y)|f^*(\KB)$ correspond for
all $\KB$, then $\P(X)$ and $\P(Y)$ must correspond: simply take $\KB =
\true$ and observe that $\P(X)|\KB) = \P(X)$ and $\P(Y)|f^*(\KB) =
\P(Y)$.
\eprf

\opro{pro:prodmeasure}
Suppose that $X_1 \times \cdots \times X_n$ is the product decomposition
on $X$ and, for each $i=1,\ldots,n$, $\KB_i$ is  a constraint 
on $X_i$, and $S_i$ is a subset of $X_i$.  Then 
\[
\band_{i=1}^n \KB_i \dentailsn_{\infp_{\P_\Pi}} \Pr(S_1 \land \ldots \land
S_n) = \prod_{i=1}^n \Pr(S_i).
\]
\eopro
\prf If $\KB_i$ is a satisfiable constraint on $\dists_{X_i}$, for $i = 1,
\ldots, n$, then there exist product measures on $X$ satisfying 
the constraints $\band_{i=1}^n \KB_i$.  These product measures are
precisely the measures in $\P_\Pi|(\band_{i=1}^n \KB_i)$.  Since each of
these measures satisfies $\Pr(S_1 \land \ldots \land S_n) = \prod_{i=1}^n
\Pr(S_i)$ by assumption, the conclusion holds in this case.  If any
constraint $\KB_i$ is not satisfiable, then the result trivially holds.
\eprf

\bigskip

\othm{thm:products}
The inference procedure $\infp_{\P_\Pi}$ is invariant under faithful
product embeddings and under permutation embeddings.
\eothm

\prf Suppose that $f$ is a faithful $X$-$Y$ product embedding, $X_1
\times \cdots \times X_n$ is the product decomposition of $X$, and 
$Y_1 \times \cdots \times Y_n$ is the product decomposition of $Y$.  To show
that ${\P_\Pi}$ is invariant under $f$, it suffices to show that
$\P_{\Pi}(X)$ and $\P_{\Pi}(Y)$ correspond under $f$.  Supposethat 
$\mu \in \P_{\Pi}(Y)$.  Then $\mu = \mu_1 \times \cdots\times \mu_n$,
where $\mu_i$ is a measure on $X_i$, $i = 1, \ldots, n$.  Moreover,
since $f$ is a product embedding, there exist $f_1, \ldots, f_n$ such
that $f = f_1 \times \cdots \times f_n$.  Let $\nu_i \in f_i^*(\mu_i)$,
for $i = 1, \ldots, n$.  It is easy to check that $\nu = \nu_1 \times
\cdots \times \nu_n \in f^*(\mu)$.  

Conversely, suppose that $\nu \in \P_{\Pi}(Y)$.  Then $\nu= \nu_1 \times
\cdots \times \nu_n$, where $\nu_i \in \dists_{Y_i}$ for $i = 1 ,
\ldots, n$.  Define $\mu \in \dists_{X_i}$ by setting $\mu_i(S) =
\nu_i(f_i(S))$.  Since $f_i$ is a faithful $X_i$-$Y_i$ embedding, is easy
to check that $\mu_i \in \dists_{X_i}$ and that $\nu_i \in
f_i^*(\mu_i)$.  Thus, $\nu \in f^*(\mu)$.
This completes the proof that $\P_\Pi$ is invariant under faithful
$X$-$Y$ product embeddings.

The argument that $\P_\Pi$ is invariant under faithful $X$-$X$
permutation embeddings is similar (and easier). We leave details to the
reader.  \eprf

\bibliographystyle{theapa}
\bibliography{z,bghk,joe,refs}
\end{document}

%% file: defn.tex
\newtheorem{THEOREM}{Theorem}[section]
\newenvironment{theorem}{\begin{THEOREM} \hspace{-.85em} {\bf :} }%
                        {\end{THEOREM}}
\newtheorem{LEMMA}[THEOREM]{Lemma}
\newenvironment{lemma}{\begin{LEMMA} \hspace{-.85em} {\bf :} }%
                      {\end{LEMMA}}
\newtheorem{COROLLARY}[THEOREM]{Corollary}
\newenvironment{corollary}{\begin{COROLLARY} \hspace{-.85em} {\bf :} }%
                          {\end{COROLLARY}}
\newtheorem{PROPOSITION}[THEOREM]{Proposition}
\newenvironment{proposition}{\begin{PROPOSITION} \hspace{-.85em} {\bf :} }%
                            {\end{PROPOSITION}}
\newtheorem{DEFINITION}[THEOREM]{Definition}
\newenvironment{definition}{\begin{DEFINITION} \hspace{-.85em} {\bf :} \rm}%
                            {\end{DEFINITION}}
\newtheorem{CLAIM}[THEOREM]{Claim}
\newenvironment{claim}{\begin{CLAIM} \hspace{-.85em} {\bf :} \rm}%
                            {\end{CLAIM}}
\newtheorem{EXAMPLE}[THEOREM]{Example}
\newenvironment{example}{\begin{EXAMPLE} \hspace{-.85em} {\bf :} \rm}%
                            {\end{EXAMPLE}}
\newtheorem{REMARK}[THEOREM]{Remark}
\newenvironment{remark}{\begin{REMARK} \hspace{-.85em} {\bf :} \rm}%
                            {\end{REMARK}}

\newcommand{\thm}{\begin{theorem}}
\newcommand{\lem}{\begin{lemma}}
\newcommand{\pro}{\begin{proposition}}
\newcommand{\dfn}{\begin{definition}}
\newcommand{\rem}{\begin{remark}}
\newcommand{\xam}{\begin{example}}
\newcommand{\cor}{\begin{corollary}}
\newcommand{\prf}{\noindent{\bf Proof:} }
\newcommand{\ethm}{\end{theorem}}
\newcommand{\elem}{\end{lemma}}
\newcommand{\epro}{\end{proposition}}
\newcommand{\edfn}{\bbox\end{definition}}
\newcommand{\erem}{\bbox\end{remark}}
\newcommand{\exam}{\bbox\end{example}}
\newcommand{\ecor}{\end{corollary}}
\newcommand{\eprf}{\bbox\vspace{0.1in}}
\newcommand{\beqn}{\begin{equation}}
\newcommand{\eeqn}{\end{equation}}

\newcommand{\bbox}{\vrule height7pt width4pt depth1pt}

\newcommand{\clm}{\begin{claim}}
\newcommand{\eclm}{\end{claim}}

\newcommand{\sat}{\models}

\newcommand{\rimp}{\Rightarrow}

\newcommand{\dimp}{\Leftrightarrow}
\newcommand{\bor}{\bigvee}
\newcommand{\band}{\bigwedge}
\newcommand{\union}{\cup}
\newcommand{\inter}{\cap}

\newcommand{\natnum}{{\sl N}}
\newcommand{\IR}{\mbox{$I\!\!R$}}

\renewcommand{\phi}{\varphi}

\newcommand{\A}{{\cal A}}

\newcommand{\D}{{\cal D}}

\newcommand{\F}{{\cal F}}

\newcommand{\I}{{\cal I}}

\renewcommand{\P}{{\cal P}}

\newcommand{\W}{{\cal W}}
\newcommand{\X}{{\cal X}}

\newcommand{\ie}{i.e.,~}

\newcommand{\ol}{\setlength{\itemsep}{0pt}\begin{enumerate}}
\newcommand{\eol}{\end{enumerate}\setlength{\itemsep}{-\parsep}}
\newcommand{\ul}{\setlength{\itemsep}{0pt}\begin{itemize}}
\newcommand{\dl}{\setlength{\itemsep}{0pt}\begin{description}}
\newcommand{\edl}{\end{description}\setlength{\itemsep}{-\parsep}}
\newcommand{\eul}{\end{itemize}\setlength{\itemsep}{-\parsep}}

\newcommand\eqdef{=_{\rm def}}
\newcommand{\true}{\mbox{{\it true}}}
\newcommand{\false}{\mbox{{\it false}}}

\newcommand{\commentout}[1]{}

\newcommand{\bi}{\begin{itemize}}
\newcommand{\ei}{\end{itemize}}
\newcommand{\be}{\begin{enumerate}}
\newcommand{\ee}{\end{enumerate}}

\newcommand{\denselist}{\itemsep 0pt\partopsep 0pt}

%% file: bghkmac.tex
\newenvironment{oldthm}[1]{\par\noindent{\bf Theorem #1:} \em \noindent}{\par}
\newenvironment{oldlem}[1]{\par\noindent{\bf Lemma #1:} \em \noindent}{\par}
\newenvironment{oldcor}[1]{\par\noindent{\bf Corollary #1:} \em \noindent}{\par}
\newenvironment{oldpro}[1]{\par\noindent{\bf Proposition #1:} \em \noindent}{\par}
\newcommand{\othm}[1]{\begin{oldthm}{\ref{#1}}}
\newcommand{\eothm}{\end{oldthm} \medskip}
\newcommand{\olem}[1]{\begin{oldlem}{\ref{#1}}}
\newcommand{\eolem}{\end{oldlem} \medskip}
\newcommand{\ocor}[1]{\begin{oldcor}{\ref{#1}}}
\newcommand{\eocor}{\end{oldcor} \medskip}
\newcommand{\opro}[1]{\begin{oldpro}{\ref{#1}}}
\newcommand{\eopro}{\end{oldpro} \medskip}

\newcommand{\KB}{{\it KB}}

\newcommand{\KBfly}{\KB_{\mbox{\scriptsize \it fly}}}

\newcommand{\bxor}[1]{\dot{\bor}}

\newcommand{\dentails}{{\;|\!\!\!\sim\;}}

\newcommand{\dentailssub}[1]{\dentails\hspace{-0.4em}_{
          \mbox{\scriptsize{\it #1}}}\;}

\newcommand{\fly}{\mbox{\it fly\/}}
\newcommand{\bird}{\mbox{\it bird\/}}